\documentclass[11pt]{article}
\usepackage{geometry}
\usepackage[pagebackref=true]{hyperref}
\usepackage{Shorthands}
\usepackage{jmlrsubfig}
\allowdisplaybreaks

\usepackage{ifthen}
\newif\ifshortver
\newcommand{\ifshort}[2]{\ifshortver#1\else#2\fi}

\title{Multi-Task Imitation Learning for Linear Dynamical Systems}
\usepackage{times}

\hypersetup{
    unicode=false,          
    pdftoolbar=true,        
    pdfmenubar=true,        
    pdffitwindow=false,     
    pdfstartview={FitH},    
    pdfnewwindow=true,      
    colorlinks=true,       
    linkcolor=blue,          
    citecolor=black,        
    filecolor=magenta,      
    urlcolor=blue,           
    breaklinks=true
}
\usepackage[round]{natbib}

\usepackage{authblk}
\newcommand*\samethanks[1][\value{footnote}]{\footnotemark[#1]}
\author[1]{Thomas T.\ Zhang\thanks{Authors contributed equally to this work.}}
\author[2]{Katie Kang\samethanks}
\author[1]{Bruce D.\ Lee\samethanks}
\author[2]{Claire Tomlin}
\author[2]{Sergey Levine}
\author[3]{Stephen Tu}
\author[1,3]{Nikolai Matni}
\affil[1]{University of Pennsylvania}
\affil[2]{University of California, Berkeley}
\affil[3]{Google Research, Brain Team}
\date{\vspace{-1.3cm}}

\begin{document}

\maketitle

\begin{abstract}%

\sloppy
We study representation learning for efficient imitation learning over linear systems.  
In particular, we consider a setting where 
learning is split into two phases: (a)
a pre-training step where a shared $k$-dimensional 
 representation is learned 
from $H$ source policies, and (b) a
target policy fine-tuning step 
where the learned representation 
is used to parameterize the policy class.
We find that the imitation gap over trajectories
generated by the learned target policy is bounded by
$\tilde{O}\left( \frac{k n_x}{HN_{\mathrm{shared}}} + \frac{k n_u}{N_{\mathrm{target}}}\right)$,
where $n_x > k$ is the state dimension,
$n_u$ is the input dimension,
$N_{\mathrm{shared}}$ denotes the total amount of data
collected for each policy during representation learning,
and $N_{\mathrm{target}}$ is the amount of target task data.
 This result formalizes the intuition that aggregating data across related tasks to learn a representation can significantly improve the sample efficiency of learning a target task. The trends suggested by this bound are corroborated in simulation. 
\end{abstract} 

\begin{center}%
    \textbf{Keywords:} Imitation learning, transfer learning, multi-task learning, representation learning
\end{center}

\section{Introduction}

Imitation learning (IL), which learns control policies by imitating expert demonstrations, has demonstrated success  across a variety of domains including self-driving cars \citep{codevilla2018end} and robotics \citep{schaal1999imitation}. However, using IL to learn a robust behavior policy may require a large amount of training data \citep{ross2011reduction}, and expert demonstrations are often expensive to collect. 
One remedy for this problem is multi-task learning: using data from other tasks (source tasks) in addition to from the task of interest (target task) to jointly learn a policy. 
We study the application of multi-task learning to IL over linear systems, and demonstrate improved sample efficiency when learning a controller via representation learning.

Our results expand on prior work that studies multi-task representation learning for supervised learning \citep{du2020few, tripuraneni2021provable},
addressing the new challenges that arise in the imitation learning setting.
%
First,
the data for IL is temporally dependent, as it is generated from a dynamical system 
$x[t+1] = f(x[t], u[t], w[t])$.
In contrast, the supervised learning setting assumes that both the train and test data are
independent and identically distributed (i.i.d.)\ from the same underlying distribution.
%
Furthermore, we are interested in the performance of the learned controller in closed-loop rather than its error on expert-controlled trajectories. 
Hence, bounds on excess risk, which corresponds to the one-step prediction error of the learned controller under the expert distribution, 
are not immediately informative for the closed-loop performance. We instead focus our analysis on the tracking error between the learned and expert policies, which requires us to account for the distribution shift between the learned and expert controllers.

We address these challenges in the setting of IL for linear systems. 
%
%
The following statement captures the benefits of multi-task representation learning
on sample complexity:
\begin{theorem}[main result, informal]
\label{thm: informal main}
Suppose that the source task controllers are sufficiently related to the target task controller. Then, the tracking error between the learned target controller and the corresponding expert is bounded with high probability by:
\begin{align*}
    \textrm{tracking error} \lesssim \frac{\textrm{rep.\ dimension } \times \textrm{state dimension}}{\textrm{\# source task datapoints}} + \frac{\textrm{rep.\ dimension } \times \textrm{input dimension}}{\textrm{\# target task datapoints}}.
\end{align*}
\end{theorem}
The first term in this bound corresponds to the error from learning a common representation, and the second term
the error in fitting the remaining weights of the target task controller. 
The key upshot of this result is that the numerator of the second term
(rep.\ dimension $\times$ input dimension)
is smaller than the number of parameters (input dimension $\times$ state dimension) in the target controller. This demonstrates an improvement in sample complexity of multi-task IL over direct IL, where the error scales as $\frac{\# \mathrm{parameters}}{\# \mathrm{data points}}$.
%
%
Furthermore, we note that the error in learning the representation decays along all axes of the data: \# of tasks $\times$ \# of trajs $\times$ traj length for source tasks, and \# of trajs $\times$ traj length for the target task.
It is non-trivial to demonstrate that the error decays with the trajectory length, and doing so requires tools that handle causally dependent data in our analysis. 

The remainder of the paper formulates the multi-task IL problem, and the assumptions required to prove \Cref{thm: informal main}.  The main contributions may be summarized as follows:
\begin{itemize}[noitemsep, nolistsep, leftmargin=*]
    \item We provide novel interpretable notions of source task overlap with the target task (\S\ref{s: problem formulation} and \S\ref{s: sample complexity}).
    \item We bound the imitation gap achieved by multi-task IL as in \Cref{thm: informal main} (\S\ref{s: sample complexity}).
    \item We empirically show the efficacy of multi-task IL when the assumptions are satisfied (\S\ref{s: numerical results}).
\end{itemize}

\subsection{Related Work}


\textbf{Multi-task imitation and reinforcement learning:} 
Multi-task RL and IL methods seek to represent policies solving different tasks with shared parameters, enabling the transfer of knowledge across related tasks \citep{teh2017distral, espeholt2018impala, hessel2018popart,singh2020scalable, deisenroth2014multitask}, and rapid test-time adaptation to new tasks \citep{finn2017maml, rakelly2019pearl, duan2016rl2, https://doi.org/10.48550/arxiv.2109.08128, Yang2021}. There also exists a body of work which theoretically analyses the sample complexity of representation learning in multi-task RL and IL \citep{https://doi.org/10.48550/arxiv.2106.08053, https://doi.org/10.48550/arxiv.2206.05900, https://doi.org/10.48550/arxiv.2205.15701, https://doi.org/10.48550/arxiv.2010.07494, https://doi.org/10.48550/arxiv.1505.06279, https://doi.org/10.48550/arxiv.2002.10544}.
While this line of work considers a more general MDP setting compared
with the linear dynamical systems we consider, the specific results
are often stated with incompatible assumptions (such as bounded states/cost functions 
and discrete action spaces), and/or do not reflect how system-theoretic properties
such as closed-loop task stability affect the final rates.

\vspace{4pt}
\noindent\textbf{Multi-task system identification and adaptive control:}
Recent work has also considered applications of multi-task learning where the dynamics change between tasks, and the goal is to perform adaptive control \citep{ harrison2018control, richards2021adaptive, richards2022control, shi2021meta, muthirayan2022meta} or dynamics forecasting \citep{wang2021meta}. Multi-task system identification \citep{modi2021joint}  and stabilization using data from related systems \citep{li2022data} have also been considered.
Our work instead studies the problem of learning to imitate different expert controllers while the system remains the same, and demonstrates bounds on the tracking error between the learned controller and its corresponding expert. 

\vspace{4pt}
\noindent\textbf{Sample complexity of multi-task learning:}
Numerous works have studied the sample efficiency gains of multi-task learning for regression and classification under various task similarity assumptions \citep{baxter1995learning, crammer2008learning, maurer2016benefit, tripuraneni2020theory, chua2021fine}. Most closely related to our results are \cite{du2020few} and \cite{tripuraneni2021provable}, both of which show multi-task representation learning sample complexity bounds in the linear regression setting in which the error from learning the representation decays with the total number of source training samples. Our work leverages these results to tackle the setting of linear imitation learning, which has the additional challenges of non-i.i.d.\ data and test time distribution shift.

\section{Problem Formulation}
\label{s: problem formulation}
\subsection{Multi-Task Imitation Learning}
\label{s: Multi-task imitation learning}

Imitation learning uses state/action pairs $(x, u) \in \R^{n_x} \times \R^{n_u}$ of expert demonstrations to learn a controller $\hat{\pi}: \mathbb{R}^{n_x} \rightarrow \mathbb{R}^{n_u}$, by matching the learned controller actions to the expert actions. In particular, if $\calD$ is the training set of expert state/action pairs, then 
    $\hat \pi \in \argmin_{\pi} \sum_{(x,u) \in \calD} \norm{ \pi(x) - u}^2.$

We are interested in the problem of \emph{multi-task} imitation learning, where we consider $H+1$ different expert controllers. We call the first $H$ controllers \textit{source controllers} and the $(H+1)^{\textrm{st}}$ controller the  \textit{target controller}. We assume that we have access to $N_1$ trajectories for each source task, and $N_2 \leq N_1$ trajectories for the target task. For simplicity, we assume all trajectories are of the same length $T$.
In particular, for each source task $h \in \curly{1, \dots, H}$, our source data consists of $\curly{\curly{(x_i^{(h)}[t], u_i^{(h)}[t])}_{t=0}^{T-1}}_{i=1}^{N_1}$, while our target data  consists of $\curly{\curly{(x_i^{(H+1)}[t], u_i^{(H+1)}[t])}_{t=0}^{T-1}}_{i=1}^{N_2}$. 
Our goal is to learn a controller which effectively imitates the target controller. However, because we only have access to a small number ($N_2$) of target expert trajectories, we leverage the $HN_1$ expert trajectories from the source controllers to accelerate the learning of the target controller.

To do so, we break our training into two stages: a pre-training stage which learns from the combined source task data, and a target training stage which only learns from the target task data. In the pre-training stage, we extract a common, low dimensional representation for the source controllers, which is used later in the target training stage. More specifically, we learn a common, low dimensional representation mapping $\hat{\phi}: \mathbb{R}^{n_x} \rightarrow \mathbb{R}^k$, where $k<n_x$ is the dimension of the representation, and linear predictors $\hat{F}^{(h)} \in \mathbb{R}^{n_u \times k}$ unique to each task:
\begin{align}
    \label{eq: pretraiing}
    \hat \phi, \hat F^{(1)}, \dots, \hat F^{(H)} \in \argmin_{\phi, F^{(1)}, \dots, F^{(H)}} \sum_{h=1}^H \sum_{i=1}^{N_1} \sum_{t=0}^{T-1} \norm{F^{(h)} \phi(x_i^{(h)}[t]) - u_i^{(h)}[t]}^2.
\end{align}
We do not address the details of solving the empirical risk minimization problem, and instead perform our analysis assuming 
\eqref{eq: pretraiing} can be solved to optimality.
Note however that \cite{tripuraneni2021provable} demonstrate in the linear regression setting that a method-of-moments-based algorithm can efficiently find approximate empirical risk minimizers.

%
Once a common representation $\hat{\phi}$ is obtained, we move on to target task training. 
During target task training, we use the common representation mapping $\hat{\phi}$ learned from the pre-training step to map the states into the lower dimensional representation, and learn an additional linear predictor $\hat{F}^{(H+1)}$ unique to the target task to model the target controller:
\begin{align}
    \label{eq: target training}
    \hat F^{(H+1)} = \argmin_{F} \sum_{i=1}^{N_2} \sum_{t=0}^{T-1} \norm{F \hat{\phi}\paren{x_i^{(H+1)}[t]} - u_i^{(H+1)}[t]}^2.
\end{align}
Since the representation $\hat{\phi}$ is fixed from  pre-training, \eqref{eq: target training} is an ordinary least squares problem. 


\subsection{System and Data Assumptions}
We focus our analysis on a linear systems setting, with state $x[t] \in \R^{n_x}$, input $u[t] \in \R^{n_u}$, and Gaussian process noise $w[t] \in \R^{n_x}$ obeying dynamics
\begin{align}
    \label{eq: linear system}
    x[t+1] = Ax[t] + Bu[t] + w[t].
\end{align}
Let each expert controller be of the form $u[t] = K^{(h)}x[t] + z[t]$, where $z[t] \in \R^{n_u}$ is Gaussian actuator noise.\footnote{As the control actions are the labels in IL, actuator noise corresponds to label noise in supervised learning. In the absence of such noise, the controller is recovered by $n_x$ linearly independent states and corresponding expert inputs.} We assume the system matrices $(A,B)$ remain the same between tasks, but the process noise covariance and the controllers may change.\footnote{This could be the case, for instance, if different controllers are designed for different levels of noise.}
In particular, we have $w^{(h)}[t] \overset{i.i.d.}{\sim} \calN(0, \Sigma_w^{(h)})$ and $z^{(h)}[t] \overset{i.i.d.}{\sim} \calN(0, \sigma_z^2 I)$ with $\Sigma_w^{(h)} \succ 0$ for all $h \in [H+1]$ and $\sigma_z^2 > 0$.


 We assume all of the expert controllers $K^{(h)}$ are stabilizing, i.e.,\ the spectral radii of $A+BK^{(h)}$ are less than one.
 Note that this implies that $(A,B)$ is stabilizable. No other assumptions on $(A,B)$ are required. 
 The state distribution of the system under each expert controller will converge to the stationary distribution $\calN(0, \Sigma_x^{(h)})$, where $\Sigma_x^{(h)}$ solves the following discrete Lyapunov equation:
\[
    \Sigma_x^{(h)} = (A+BK^{(h)})\Sigma_x^{(h)} (A+BK^{(h)})^\top +\sigma_z^2 B B^\top + \Sigma_w^{(h)}.
\]
For simplicity, we assume that the initial states of the expert demonstrations in our datasets are sampled $x_i^{(h)}[0] \overset{i.i.d.}{\sim} \calN(0, \Sigma_x^{(h)})$. Thus at all times, the marginal state distributions of the expert demonstrations are equal to $\calN(0, \Sigma_x^{(h)})$.\footnote{This assumption is not restrictive, as stable systems exponentially converge to stationarity.}

Finally, we assume that the expert controllers share a low dimensional representation. 
Specifically, there exists some $\Phi_\star \in \R^{k \times n_x}$ with $2k \leq n_x$\footnote{While this assumption is more stringent than the intuitive $k<n_x$  assumption, it arises from the fact that the residual of the stacked source controllers may be of rank $2k$.} 
and weights $F^{(1)}_\star, F^{(2)}_\star, \dots, F^{(H+1)}_\star \in \R^{n_u \times k}$ such that for all $h\in[H+1]$, $K^{(h)} = F_\star^{(h)} \Phi_\star$, and the action taken at time $t$ for trajectory $i$ is:\footnote{An example of a setting where expert controllers satisfy this assumption is when the system has high dimensional states which exhibit low dimensional structure, e.g.\ when $A$ and $B$ can be decomposed into $A = \Phi_\star^{\dagger} \tilde{A} \Phi_\star$ and $B = \Phi_\star^{\dagger} \tilde{B}$, where $\tilde{A} \in \mathbb{R}^{k \times k}$ and $\tilde{B} \in \mathbb{R}^{k \times n_u}$. Here, linear policies $K$ which optimize some objective in terms of the low dimensional features of the system can be decomposed into $K = \tilde{K} \Phi_\star$, where $\tilde{K} \in \mathbb{R}^{n_u \times k}$, mirroring the assumptions of our expert controllers. We provide a concrete example in Section \ref{s: numerical results}. }
\begin{equation*}
    u_i^{(h)}[t] = F^{(h)}_\star \Phi_\star x_i^{(h)}[t] + z_i^{(h)}[t].
\end{equation*}
Under this assumption, the learned common representation $\hat \phi$ in \Cref{s: Multi-task imitation learning} can be restricted to linear representations, i.e.,\ $\hat \phi(x) = \hat \Phi x$, where $\hat \Phi \in \R^{k \times n_x}$. Note that solving Problem \eqref{eq: target training} with $\hat \Phi$ fixed involves solving for only $k n_u$ parameters, which is smaller than the $n_u n_x$ unknown parameters when learning from scratch. In particular, by representing the controller as $F^{(H+1)} \Phi$, we have $k (n_u +n_x)$ unknown parameters: $k n_x$ of the parameters are, however, learned using the source task data, leaving only $k n_u$ parameters to learn with target task data. 

\subsection{Notation} 
The Euclidean norm of a vector $x$ is denoted $\norm{x}$. For a matrix $A$, the spectral norm is denoted $\norm{A}$, and the Frobenius norm is denoted $\norm{A}_F$. The spectral radius of a square matrix is denoted $\rho(A)$. We use $\dagger$ to denote the Moore-Penrose pseudo-inverse.  For a square matrix $A$ with $\rho(A) < 1$, define $\calJ(A) = \sum_{t \geq 0} \norm{A^t} < \infty$. A symmetric, positive semi-definite (psd) matrix $A = A^\top$ is denoted $A \succeq 0$.  Similarly $A \succeq B$ denotes that $A-B$ is positive semidefinite. 
The condition number of a positive definite matrix $A$ is denoted $\kappa(A) = \frac{\lambda_{\max}(A)}{\lambda_{\min}(A)}$, where $\lambda_{\max}$ and $\lambda_{\min}$ denote the maximum and minimum eigenvalues, respectively. Similarly, $\sigma_i(A)$ denote the singular values of $A$. We denote the normal distribution with mean $\mu$ and covariance $\Sigma$ by  $\calN(\mu, \Sigma)$. We use standard $\calO(\cdot)$, $\Theta(\cdot)$ and $\Omega(\cdot)$ to omit universal constant factors, and $\tilde \calO(\cdot), \tilde \Theta(\cdot)$ and $\tilde \Omega(\cdot)$ to also omit polylog factors. We also use $a \lesssim b$ to denote $a = O(b)$.
We use the indexing shorthand $[K] := \curly{1,\dots,K}$. For a given task $h \in [H+1]$, the matrix of stacked states is defined as 
\begin{align}
    \label{eq:stacked xs}
    \Xdata^{(h)} &= \bmat{x_1^{(h)}[0] & \dots & x_1^{(h)}[T-1] & \dots & x_{N_1}^{(h)}[0] & \dots & x_{N_1}^{(h)}[T-1]}^\top \in \R^{N_1 T \times n_x}.
\end{align}
Lastly, let $\bar \lambda = \max_{1 \leq h \leq H} \lambda_{\max}(\Sigma_x^{(h)})$ and $\underline \lambda = \min_{1 \leq h \leq H} \lambda_{\min}(\Sigma_x^{(h)})$. 
\section{Sample Complexity of Multi-Task Imitation Learning}
\label{s: sample complexity}
\sloppy

\noindent In order to derive any useful information from source tasks for a downstream task, the source tasks must satisfy some notion of \textit{task diversity} that sufficiently covers the downstream task. To that end, we introduce the following notions of source tasks covering the target task.
\begin{definition}[target task covariance coverage \citep{du2020few}]
\label{def: target task covariance coverage}
Define the constant $c$ as:
\begin{align}
    \label{eq: target task covariance coverage}
    c := \min_{h \in [H]} \lambda_{\min}( (\Sigma_x^{(H+1)})^{-1/2}  \Sigma_x^{(h)} (\Sigma_x^{(H+1)})^{-1/2} ).
\end{align}
Note that $c$ is well-defined and positive
by our assumption that $\Sigma_w^{(h)} \succ 0$ for all $h \in [H + 1]$.   
\end{definition}
\Cref{def: target task covariance coverage} captures the degree to which the closed-loop distribution of states for each source task aligns with that of the target task.
We then introduce the following notion of task similarity between the source and target task weights, which generalizes the well-conditioning assumptions in \cite{du2020few} and \cite{tripuraneni2021provable}.
\begin{assumption}[diverse source controllers] 
    \label{as: diverse source controllers}
We assume the target task weights $F_\star^{(H+1)}$ and the  source task weights $F_\star^{(1)}, \dots, F_\star^{(H)}$  satisfy 
\begin{align}\label{eq: diverse source controllers}
    \norm{F_\star^{(H+1)} \bmat{F_\star^{(1)} \\ \vdots \\ F_{\star}^{(H)}}^\dagger}^2 \leq \calO\paren{\frac{1}{H}}.
\end{align}
\end{assumption}
\Cref{as: diverse source controllers} states that the alignment and loadings of the singular spaces between the stacked source task weights and target task weights closely match along the low-dimensional representation dimension. For example, if $F_\star^{(h)} = F_\star^{(H+1)}$ for each $h \in [H]$, the RHS of~\eqref{eq: diverse source controllers} is $1/H$.
We note that this assumption subsumes and is more geometrically informative than a direct bound on the ratio of singular values, e.g.\ $$\sigma_{\max}^2(F_\star^{(H+1)})/\sigma_k^2\paren{\bmat{F_\star^{(1)} \\ \vdots \\ F_{\star}^{(H)}}} \leq \calO(1/H),$$ which would follow by naively extending the well-conditioning assumptions in \cite{du2020few} and \cite{tripuraneni2021provable}. Notably, such a condition might not be satisfied even if $F_\star^{(h)} = F_\star^{(H+1)}$, $\forall h \in [H]$, e.g.,\ if $F_\star^{(H+1)}$ is rank-deficient. 
\subsection{Excess Risk Bound: Generalization Along Expert Target Task Trajectories}

First we show that learning controllers through multi-task representation learning leads to favorable generalization bounds on the excess risk of the learned controller inputs on the expert target task state distribution, analogous to the bounds on multi-task linear regression in \cite{du2020few, tripuraneni2021provable}. However, a key complicating factor in our setting is the fact that the input noise $z^{(h)}[t]$ enters the process, and thus the data $x^{(h)}[t]$ is causally dependent on the ``label noise''. In order to overcome this issue and preserve our statistical gains along time $T$, we leverage the theory of self-normalized martingales, in particular generalizing tools from \cite{abbasi2011online} to the matrix-valued setting. The full argument is detailed in \ifshort{\cite{zhang2022multitask}}{\Cref{appendix: covariance concentration} and \Cref{appendix: data guarantees}}. This culminates in the following target task excess risk bound.
\begin{restatable}[target task excess risk bound]{theorem}{TargetTaskRiskBound}
\label{thm: target task excess risk bound}
    Given $\delta \in (0,1)$, suppose that  
    \begin{align*}
        N_1 T &\gtrsim \max_{h \in [H]} \calJ\paren{A + B K^{(h)}}^2 \kappa\paren{\Sigma_x^{(h)}}(n_x + \log(H/\delta)), \\
        N_2 T &\gtrsim \calJ\paren{A + B K^{(H+1)}}^2 \kappa\paren{\Sigma_x^{(H+1)}}(k + \log(1/\delta)).
    \end{align*}
    Define $\calP_{0:T-1}^{(H+1)}$ as the distribution over target task trajectories $(x^{(H+1)}[0], \cdots, x^{(H+1)}[T-1])$. Then with probability at least $1-\delta$, the excess risk of the learned representation $\hat{\Phi}$ and target task weights $\hat F^{(H+1)}$ is bounded by
    \begin{align}
         \ER(\hat \Phi, \hat F^{(H+1)}) &:= \frac{1}{2 T} \Ex_{\calP_{0:T-1}^{(H+1)}} \brac{\sum_{t=0}^{T-1} \norm{(F_\star^{(H+1)} \Phi_\star - \hat F^{(H+1)} \hat \Phi) x^{(H+1)}[t]}^2} \nonumber \\
         &\lesssim  \sigma_z^2 \paren{\frac{k n_x \log\paren{N_1 T \frac{\bar \lambda}{\underline{\lambda}}} }{ cN_1 T H} + \frac{kn_u +\log(\frac{1}{\delta})}{N_2 T}}. \label{eq: excess risk bound}
    \end{align}
\end{restatable}

Note that when we are operating in the setting where we have much more source data than target data, the second term limits the excess risk bound in \eqref{eq: excess risk bound}. The second term scales with $k n_u$, which is smaller than the number of total parameters in the controller $n_u n_x$, or $k(n_u + n_x)$ under the assumption of a low rank (rank-$k$) controller. Therefore, the benefit of multi-task learning exhibited by this bound is most clear in the setting of underactuation, i.e.,\ when $n_u \leq n_x$. It should also be noted that the quantity $k n_x$ in the numerator of the first term will only be smaller than the number of source controller parameters ($n_x n_u H$) if $k$ is much smaller than $n_u H$. This is reasonable, because if $k \geq n_u H$, an optimal representation could simply contain all of the source task controllers. 

\subsection{Closed-Loop Guarantees: Tackling Distribution Shift}
We show that using multi-task representation learning leads to favorable generalization bounds of the performance of the learned target controller in closed-loop. 
As we are studying the pure offline imitation learning (``behavioral cloning'') setting, we do not assume that the expert controllers are optimizing any particular objective.
Therefore, to quantify the performance of the controller, we bound the deviation of states generated by the learned and expert target controller run in closed-loop, i.e., the \textit{tracking error}, which implies general expected-cost bounds.

In order to transfer a bound on the excess risk of the target task $\ER(\hat \Phi, \hat F^{(H+1)})$ into a bound on the tracking error, we must account for the fundamental distribution shift between the expert trajectories seen during training and the trajectories generated by running the learned controller in closed-loop. We leverage the recent framework of \cite{pfrommer2022tasil} to bound the tracking error, making the necessary modifications to handle stochasticity. Our bound formalizes the notion that ``low training error implies low test error,'' even under the aforementioned distribution shift. A detailed exposition can be found in \ifshort{\cite{zhang2022multitask}}{\Cref{appendix: bounding imitation gap}}.

Let us define the following coupling of the states of the expert versus learned target task closed-loop systems: given a learned controller $\hat K = \hat F^{(H+1)} \hat \Phi$ from solving the pre-training and fine-tuning optimization problems \eqref{eq: pretraiing} and \eqref{eq: target training}, for a realization of process randomness $x[0] \sim \calN(0, \Sigma_x^{(H+1)})$ and $z[t] \overset{\mathrm{i.i.d.}}{\sim} \calN(0, \sigma_z^2 I)$, $w[t] \overset{\mathrm{i.i.d.}}{\sim} \calN(0, \Sigma_w^{(H+1)})$ for $t = 0, \dots, T-1$, we write
\begin{align*}
    x_\star[t+1] &= (A + BK^{(H+1)}) x_\star[t] + Bz[t] + w[t], \quad x_\star[0] = x[0], \\
    \hat x[t+1] &= (A + B\hat K)\hat x[t] + Bz[t] + w[t],  \quad \hat x[0] = x[0].
\end{align*}
Thus $\hat x[t]$ and $x_\star[t]$ are the states visited by the learned and expert target task systems with the \emph{same}
draw of process randomness. We show a high probability bound on the closed-loop tracking error $\norm{x_\star[t] - \hat x[t]}$ 
that scales with the excess risk of the learned controller. Denote by $\calP^\star_{1:T}$ and $\hat\calP_{1:T}$ the distributions of trajectories $\curly{x_\star[t]}_{t=1}^{T}$ and $\curly{\hat x[t]}_{t=1}^{T}$.

\begin{restatable}[Target task tracking error bound]{theorem}{TargetImitationErrorBound} \label{thm: final imitation gap bound short}
Let 
$(\hat\Phi, \hat F^{(H+1)})$ denote the learned representation and target task weights,
and $\mathrm{ER}(\hat \Phi, \hat F^{(H+1)})$ denote the corresponding excess risk. Define $A_{\mathsf{cl}} := A+BK^{(H+1)}$. Assume that the excess risk satisfies:
\begin{equation}
     \mathrm{ER}(\hat \Phi, \hat F^{(H+1)}) \lesssim \frac{\lambda_{\min}\paren{\Sigma_x^{(H+1)}}}{\calJ\paren{A_{\mathsf{cl}}}^2\norm{B}^2}. \label{eq:ER_requirement}
\end{equation}
Then with probability greater than $1- \delta$, for a new target task trajectory sampled with process randomness $x[0] \sim \calN(0, \Sigma_x^{(H+1)})$ and $z[t] \overset{\mathrm{i.i.d.}}{\sim} \calN(0, \sigma_z^2 I)$, $w[t] \overset{\mathrm{i.i.d.}}{\sim} \calN(0,\Sigma_w^{(H+1)})$ for $t = 0, \dots, T-1$, the tracking error satisfies
\begin{align}
    \label{eq: Thm 3.2 high prob bound}
    \max_{1 \leq t \leq T} \norm{\hat x[t] - x_\star[t]}^2 &\lesssim \calJ\paren{A_{\mathsf{cl}}}^2\norm{B}^2 \log\paren{\frac{T}{\delta}} \mathrm{ER}(\hat{\Phi}, \hat F^{(H+1)}).
\end{align}
Furthermore, for any cost function $h(\cdot)$ that is $L$-Lipschitz with respect to the trajectory-wise metric $d\paren{\vec x_{1:T}, \vec y_{1:T}} = \max_{1 \leq t \leq T} \norm{x[t] - y[t]}$, we have the following bound on the expected cost gap
\begin{align}\label{eq: Thm 3.2 expectation bound}
    \abs{\Ex_{\hat\calP_{1:T}}\brac{h(\hat{\vec x}_{1:T})} - \Ex_{\calP^\star_{1:T}}\brac{h(\vec x^\star_{1:T})}} &\lesssim L\calJ(A_{\mathsf{cl}})\norm{B} \sqrt{\log{T}} \sqrt{\mathrm{ER}(\hat{\Phi}, \hat F^{(H+1)})} 
\end{align}
\end{restatable}
By invoking the bound on the excess risk from \Cref{thm: target task excess risk bound}, condition \eqref{eq:ER_requirement} is satisfied with probability at least $1-\delta'$ if we have sufficiently many samples $H$, $T$, $N_1$, $N_2$ such that
\[
    \sigma_z^2 \paren{\frac{k n_x \log\paren{N_1 T \frac{\bar \lambda}{\underline{\lambda}}} }{ cN_1 T H} + \frac{kn_u +\log(\frac{1}{\delta'})}{N_2 T}} \lesssim \frac{\lambda_{\min}\paren{\Sigma_x^{(H+1)}}}{\calJ\paren{A_{\mathsf{cl}}}^2\norm{B}^2}.
\]
The bound on excess risk from \Cref{thm: target task excess risk bound} may also be substituted into the tracking error bound in \eqref{eq: Thm 3.2 high prob bound} to find that with probability at least $1-\delta - \delta'$, the tracking error satisfies
\[
    \max_{1 \leq t \leq T} \norm{\hat x[t] - x_\star[t]}^2 \lesssim \calJ\paren{A_{\mathsf{cl}}}^2\norm{B}^2 \log\paren{\frac{T}{\delta}} \sigma_z^2 \paren{\frac{k n_x \log\paren{N_1 T \frac{\bar \lambda}{\underline{\lambda}}} }{ cN_1 T H} + \frac{kn_u +\log(\frac{1}{\delta'})}{N_2 T}}.
\]
The above inequality provides the informal statement of the main result in \Cref{thm: informal main} by hiding $\log$ terms as well as the terms dependent on system parameters. 
A bound for the expected cost gap $\abs{\Ex_{\hat\calP_{1:T}}\brac{h(\hat{\vec x}_{1:T})} - \Ex_{\calP^\star_{1:T}}\brac{h(\vec x^\star_{1:T})}}$ can be similarly instantiated. 
\begin{remark}\label{remark: spectral radius dependence}
    The dependence of the tracking error bound in \eqref{eq: Thm 3.2 high prob bound} on the stability of the target-task closed-loop system through $\calJ(A_{\mathsf{cl}})$ is tight (see  \ifshort{\cite{zhang2022multitask}}{\Cref{appendix: bounding imitation gap}}). Intuitively, less stable systems exacerbate the input errors from the learned controller.
\end{remark}

\begin{remark}\label{remark: Lipschitz costs}
    Some immediate examples of $h(\cdot)$ include LQR state costs $h(\vec x_{1:T}) = \max_t \norm{Q^{1/2} x[t]}$ and regularized tracking costs $h(\vec x_{1:T}) = \max_t \norm{x[t] - x_{\mathrm{goal}}[t]} + \lambda \norm{Rx[t]}$. Since $\frac{1}{T}\sum_{t=1}^{T} \norm{x[t] - y[t]} \leq \max_{1 \leq t \leq T} \norm{x[t] - y[t]}$, \eqref{eq: Thm 3.2 expectation bound} holds with no modification for time-averaged costs $h(\cdot)$. Bounds on the full LQR cost $h\paren{\paren{\vec x_{1:T}, K}}
    := \max_{1 \leq t \leq T} \norm{\bmat{Q^{1/2} \\ R^{1/2} K} x[t]}$ can be similarly derived, and are detailed in \Cref{appendix: LQR bounds}.
\end{remark}

\section{Numerical Results}
\label{s: numerical results}

We consider a simple system with $n_x=4$ and $n_u=2$ from \cite{hong2021lecture}. 
In particular, let
\begin{align*}
    x[t+1] = \bmat{.99 & .03 & -.02 & -.32 \\ .01 & .47 & 4.7 & .00 \\ .02 & -.06 & .40 & .00 \\ .01 & -.04 & .72 & .99} x[t] + \bmat{.01 & .99 \\ -3.44 & 1.66\\ -.83 & .44 \\ -.47 & .25} u[t] =: A x[t] +  B u[t]. 
\end{align*}
We generate a collection of stabilizing controllers $K^{(1)}$, $K^{(2)}, \dots, K^{(H+1)}$ as LQR controllers with different cost matrices. Specifically, let $R=I_2$, and $Q^{(h)} = \alpha^{(h)} I_4$ for $\alpha^{(h)} \in \texttt{logspace}(-2,2,H+1)$, where $H=9$. The controllers $K^{(h)}$ are then given by $K^{(h)} = -(B^\top P^{(h)} B + R)^{-1} B^\top P^{(h)} A$, where $P^{(h)}$ solves the following Discrete Algebraic Riccati equation: $P^{(h)} =A^\top P^{(h)} A + A^\top P^{(h)} B (B^\top P^{(h)} B + R)^{-1} B^\top P^{(h)} A + Q^{(h)}$.

Next, assume that rather than directly observing the state, we obtain a high dimensional observation given by an injective linear function of the state: such an observation model can be viewed as a linear ``perceptual sensor'' or camera.
In particular, we suppose that $y_t = G x_t$, where $G \in \R^{50 \times 4}$. For simplicity, we select the elements of $G$ i.i.d.\ from $\calN(0,1)$, which ensures that $G$ is injective almost surely.
The dynamics of the observations may be written $
    y[t+1] = GA x_t + GBu[t] = GAG^{\dagger} y[t] + G B u[t],$
with the input $u[t] = K^{(h)} x[t] = K^{(h)} G^{\dagger} y[t]$. Define $\bar A = G A G^{\dagger}$ and $\bar B = G B$, and $\bar K^{(h)} = K^{(h)} G^\dagger$. Consider the dynamics in the face of process noise $w[t] \overset{i.i.d.}{\sim}\calN(0, I_{50})$, along with inputs corrupted by noise $z[t]\overset{i.i.d.}{\sim}\calN(0,I_2)$:
\begin{align}
    \label{eq: lifted}
    y[t+1] &= (\bar A + \bar B \bar K^{(h)}) y[t] + \bar B z[t]+ w[t], \quad u[t] = \bar K^{(h)} y[t] + z[t].
\end{align}

For the first $H$ controllers, we collect $N_{1}$ trajectories of length $T=20$ to get the pairs $\curly{\curly{\curly{(y_i^h[t], u_i^h[t])}_{t=0}^{T-1}}_{i=1}^{N_1}}_{h=1}^H$. For the last controller, we collect $N_2$ length $T=20$ trajectories to get the dataset $\curly{\curly{(y_i^{H+1}[t], u_i^{H+1}[t])}_{t=0}^{T-1}}_{i=1}^{N_2}$. Our goal is to learn the controller $\bar K^{(H+1)}$ from the collected state measurements and inputs. We compare the following ways of doing so:
\begin{figure}[t]
    \centering
    \subfigure[Tracking Error]
    {\label{fig:target imitation}\includegraphics[width=0.35\textwidth]{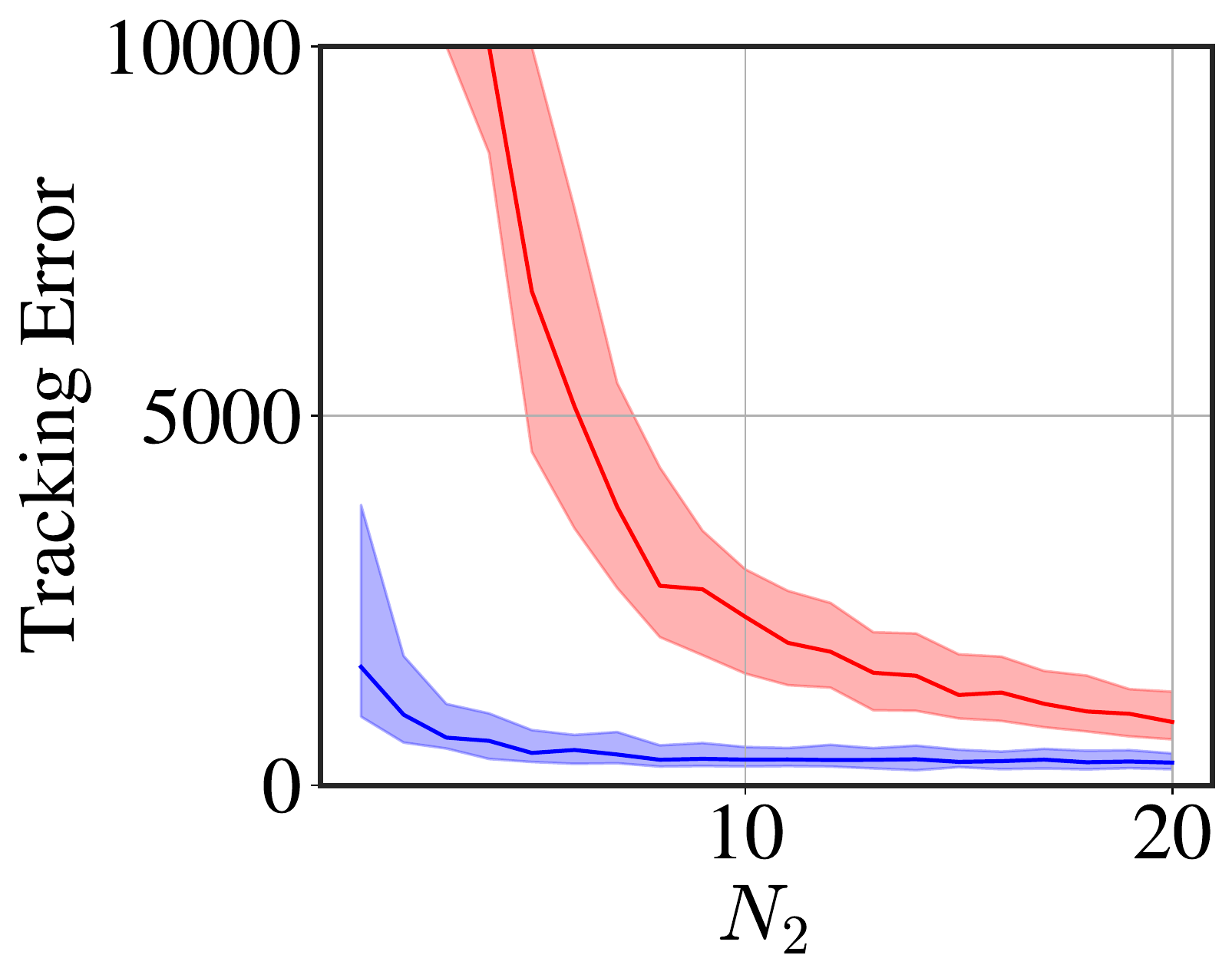}}
    \subfigure[Parameter Error]
    {\label{fig:target parameter}\includegraphics[width=0.3075\textwidth]{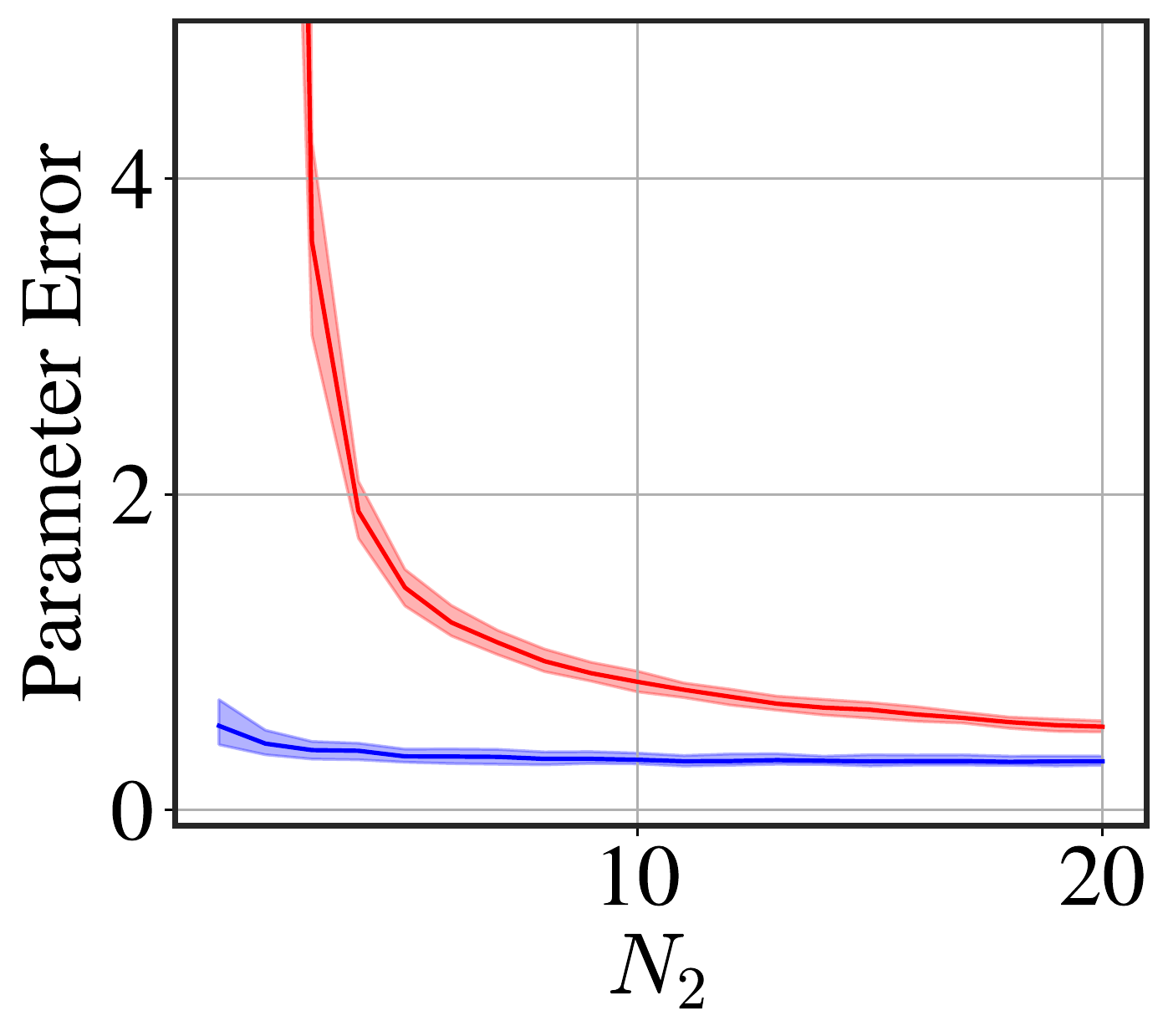}}
    \subfigure[Percent Stable]{\label{fig:target stable}\includegraphics[width=0.325\textwidth]{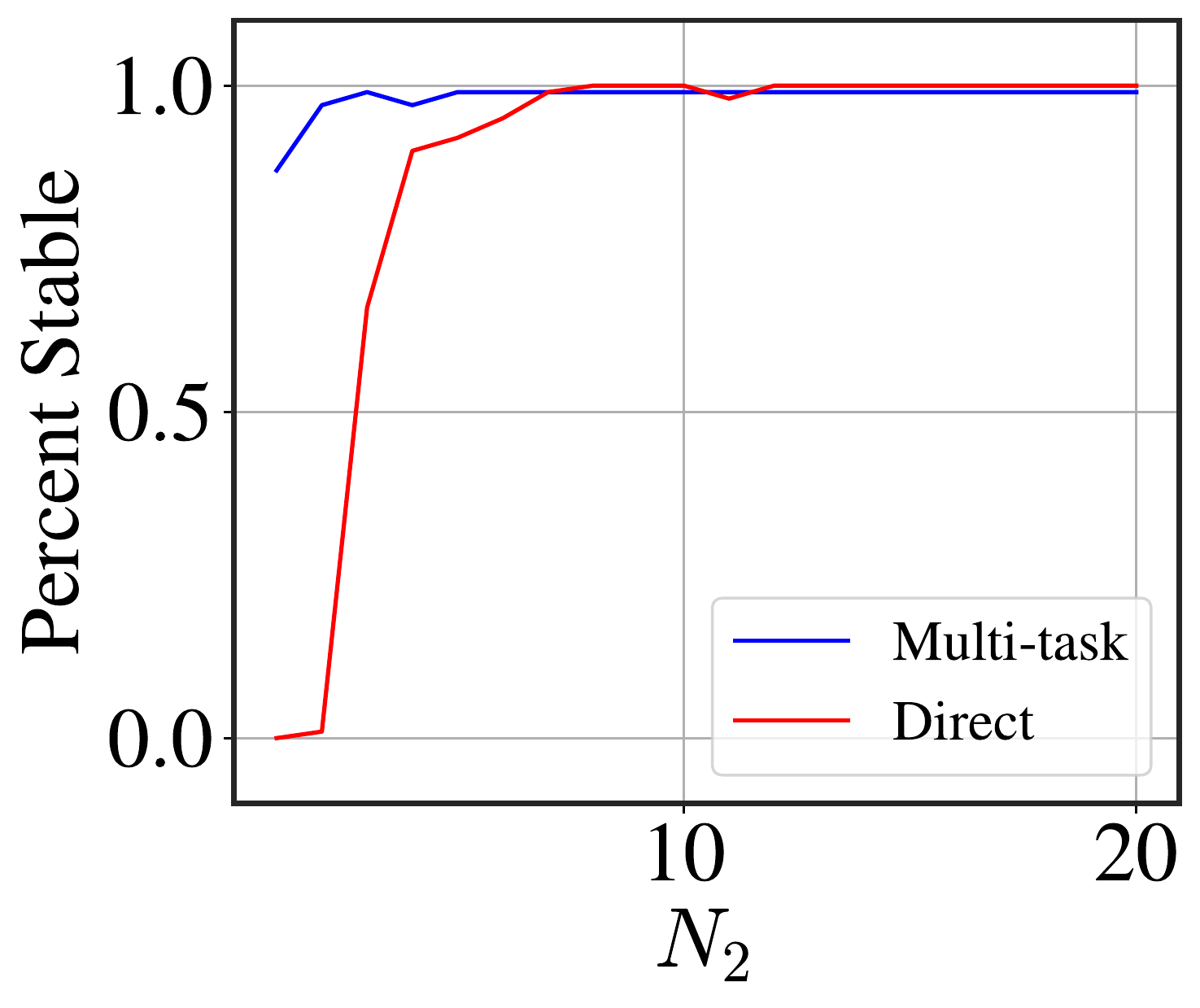}}
    \vspace{-6pt}
    \caption{We plot the tracking error between trajectories from the expert and learned controllers, $\underset{{1 \leq t \leq T_{\textrm{test}}}}{\max} \norm{\hat y[t] - y_\star[t]}^2$, the parameter error, $\paren{\norm{\hat F^{(H+1)} \hat \Phi - \bar K^{(H+1)}}_F}$, and the percent of stable closed-loop systems for varying amounts of target task data to compare multi-task IL to directly learning the controller from target task data only. All metrics are plotted with respect to the lifted system in Equation \eqref{eq: lifted}. Multi-task IL demonstrates a significant benefit over direct IL in all metrics, especially when there is limited target task data. }
    \ifshort{\vspace{-18 pt}}{}
    \label{fig:target training}
    \end{figure}
\begin{itemize}[noitemsep, nolistsep, leftmargin=*]
     \item \textbf{Multi-task Imitation Learning:} We observe that the data generating mechanism ensures the existence of a low dimensional representation. In particular, one possible $\Phi_\star$ is $G^\dagger$. Therefore, the stage is set for the two step approach outlined in Section~\ref{s: problem formulation}.  In particular, we assume that the true underlying state dimension is known, and we set the low dimensional representation dimension to $k=4$, and jointly optimize over $\Phi$, $F^{(1)}, \dots, F^{(H)}$ in Problem \eqref{eq: pretraiing}. We approximately solve this problem with $10000$ steps of alternating gradient descent using the \texttt{adam} optimizer \citep{kingma2014adam} in \texttt{optax} \citep{deepmind2020jax} with a learning rate of $0.0001$.
    The learned representation is then fixed, and the target training data is used to optimize $F^{(H+1)}$. 
    \item \textbf{Direct Imitation Learning:} We compare multi-task learning to direct learning, which does not leverage the source data. In particular, given the target data, direct learning solves the problem $\minimize_{F^{(H+1)}} \sum_{i=1}^{N_1} \sum_{t=0}^{T-1} \norm{F^{(H+1)} y_i^{(H+1)}[t] - u_i^{(H+1)}[t]}^2$ \footnote{Note that another baseline leverages the fact that a $k$ dimensional representation of the state exists, and learns it using target task data only by solving $\minimize_{F^{(H+1)}, \Phi} \sum_{i=1}^{N_1} \sum_{t=0}^{T-1} \norm{F^{(H+1)} \Phi y^{(H+1)}_i[t] - u_i^{(H+1)}[t]}^2$. For the current example, however, $n_u < k$, so this approach is less efficient than the direct learning approach proposed.}. 
    In this setting, we let $\hat \Phi = I_{50}.$
\end{itemize}

Note that a $2 \times 50$ controller has $n_u \times n_x = 100$ parameters to learn from the target data. Meanwhile, multi-task imitation learning needs to learn a total of $k \times n_u + k \times n_x = 208$ parameters for the target controller, but the $k \times n_u$ parameters are learned using source task data. This leaves only $8$ parameters to learn using target task data. 
    
\Cref{fig:target training} plots three metrics that provide insight into the efficacy of these approaches: the imitation gap given by  $\max_{1 \leq t \leq T_{\textrm{test}}} \norm{\hat y[t] - y_\star[t]}^2$ for length $T_{\textrm{test}} = 100$ observation trajectories $\hat y$ and $y_\star$ rolled out under the learned controller and expert controller, respectively, with the same noise realizations; the parameter error, $\norm{\hat F^{(H+1)} \hat \Phi - \bar K^{(H+1)}}_F$; and the percentage of trials where the learned controller is stabilizing. The trials are over ten realizations of $G$, as well as ten realizations of the noise, for a total of $100$ trials. For each trial, $N_1=10$, while $N_2$ sweeps values in $[20]$. The medians for the imitation gap and parameter error are shown, with the $20\%-80\%$ quantiles shaded. 

 \begin{figure}[t]
    \centering
    \subfigure[Tracking Error]
    {\label{fig:source imitation}\includegraphics[width=0.35\textwidth]{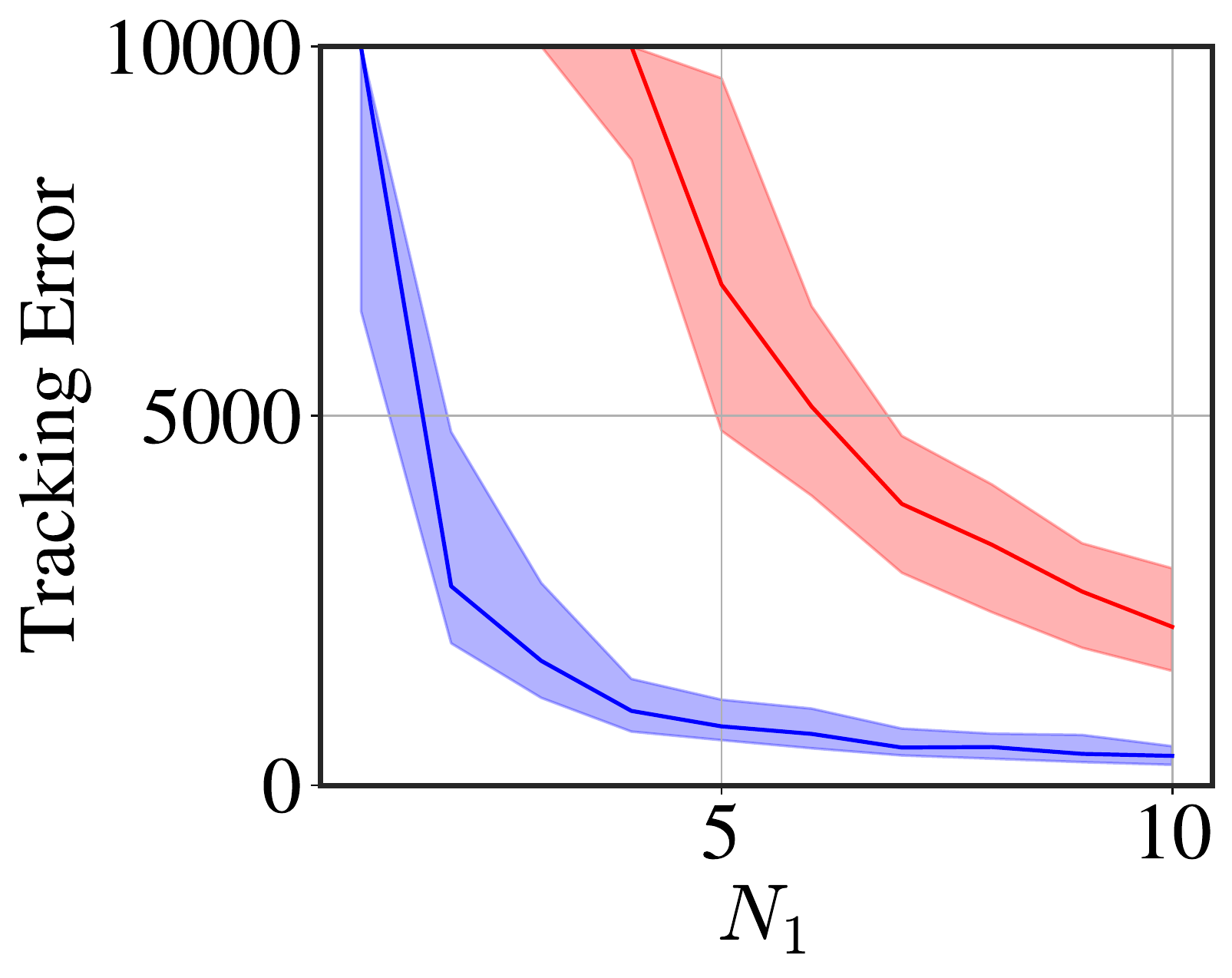}}
    \subfigure[Parameter Error]
    {\label{fig:source parameter}\includegraphics[width=0.3075\textwidth]{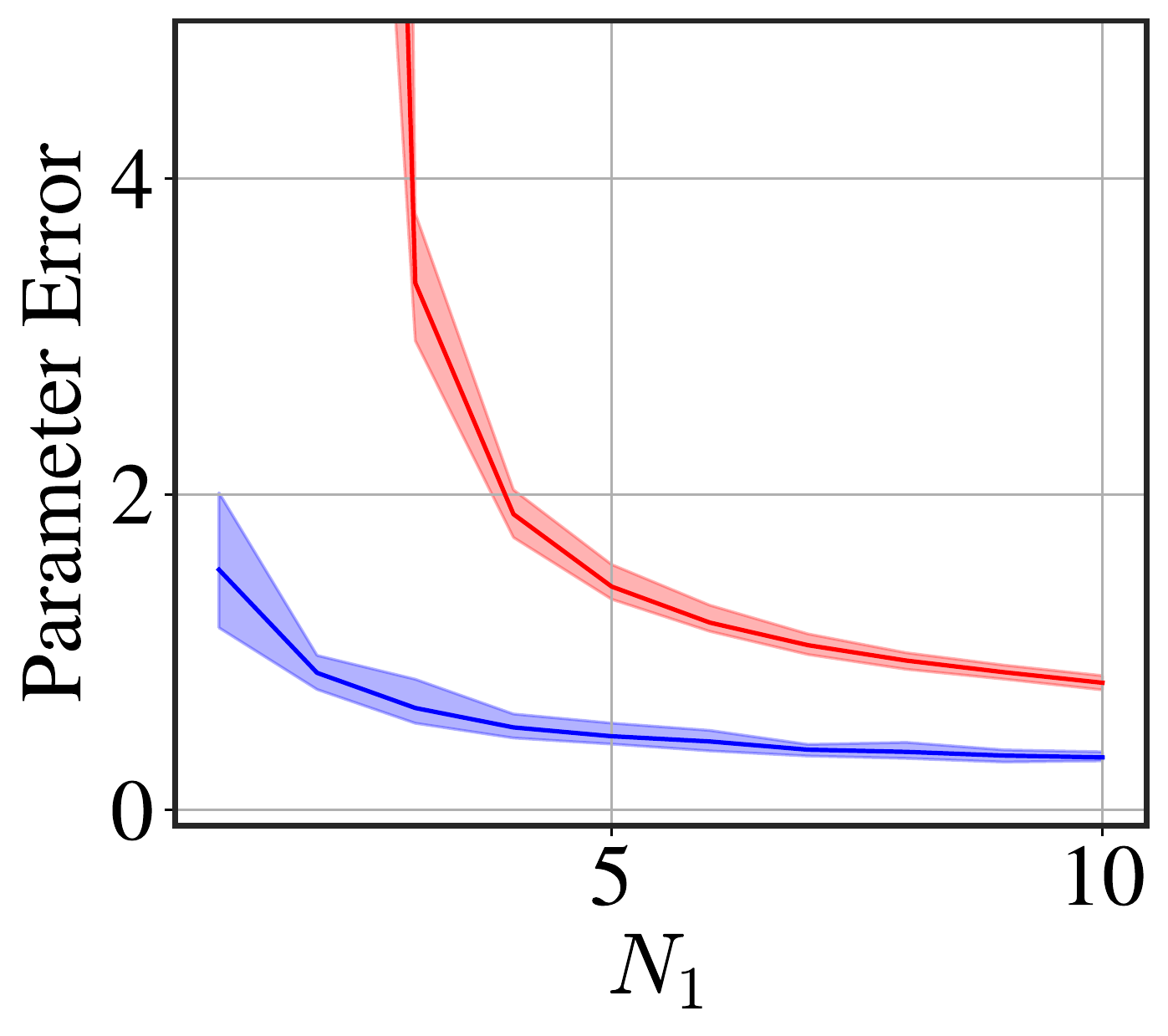}}
    \subfigure[Percent Stable]{\label{fig:source stable}\includegraphics[width=0.325\textwidth]{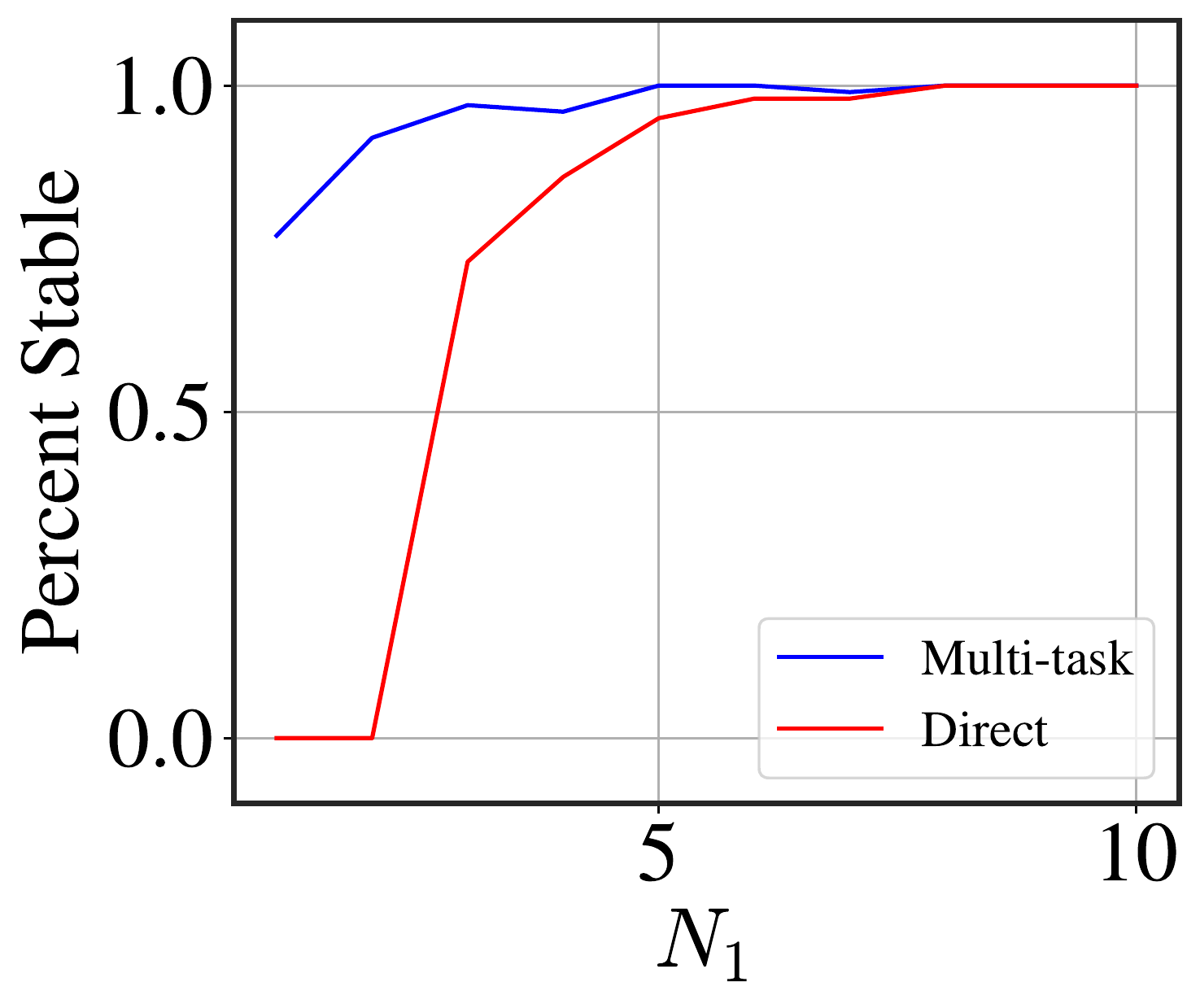}}
    \vspace{-6pt}
    \caption{We plot the tracking error between trajectories from the expert and learned controllers, $\underset{{1 \leq t \leq T_{\textrm{test}}}}{\max} \norm{\hat y[t] - y_\star[t]}^2$, the parameter error, $\paren{\norm{\hat F^{(H+1)} \hat \Phi - \bar K^{(H+1)}}_F}$, and the percent of stable closed-loop systems for varying amounts of target task data to compare multi-task IL to directly learning the controller from target task data only.
    Similar to the setting of multi-task IL for transfer to a new task, leveraging all source data to learn the controller for a single source task provides a significant benefit in all three metrics over direct IL. }
    \ifshort{\vspace{-18 pt}}{}
    \label{fig:multitask training}
\end{figure}
In Figure~\ref{fig:multitask training}, we additionally plot these metrics on one of the $H$ source training tasks (arbitrarily chosen as $h=7$) for varying amounts of training data to demonstrate the efficacy of the approach for multi-task learning. Here, $N_1$ ranges from $1$ to $10$. We compare the controller $\hat F^{(h)} \hat \Phi$ resulting from the shared training in Problem \eqref{eq: pretraiing} with the controller from directly training a controller on this task without leveraging source task data. We note that our theoretical results, with mild modification, also support the efficacy of this simultaneous training of a representation and task weights.

\section{Conclusion and Future Work}


\label{s: conclusion}

In this work, we study the sample complexity of multi-task imitation learning for linear systems. We find that if the different sets of expert demonstrations share a low dimensional representation, and the demonstrations are sufficiently diverse, then doing multi-task representation learning will lead to a smaller tracking error when deploying the learned policy in closed-loop, compared to learning a policy only from target task data. Our results are a first step towards understanding how the performance of a controller trained on multi-task data relates to the characteristics of the multi-task training data and the system being controlled. Some exciting directions for future work would be to extend our analysis to nonlinear systems and nonlinear representation functions, as well as other types of learning algorithms such as model-based and model-free RL.

\section*{Acknowledgements}
Katie Kang is supported by a NSF Graduate Research Fellowship. 
Bruce D. Lee is supported by the DoD through a National Defense Science \& Engineering Fellowship. 
Sergey Levine is supported in part by the DARPA Assured Autonomy program.
Nikolai Matni and Thomas Zhang are supported in part by NSF awards CPS-2038873, CAREER award ECCS-2045834, and a Google Research Scholar award.

\bibliographystyle{unsrtnat}
\bibliography{refs}

\clearpage
\appendix

\paragraph{Additional Notation}
The set of $n \times m$ matrices $(n \geq m)$ with orthonormal columns is denoted by $\calO_{n,m}$. The set of $d$ dimensional unit vectors is denoted by $\calS^{d-1}$. We let $P_A = A(A^\top A)^\dagger A^\top$ denote the projection onto $\Span(A)$, and let $P_A^\perp = I - P_A$ denote the projection onto $\Span(A)^\perp$. The Kronecker product of a matrix $A$ with a matrix $B$ is denoted $A \otimes B$. The vectorization of a matrix $A$ with columns $A_1, \dots, A_n$ is denoted $\vecop(A) = \bmat{A_1^\top & A_2^\top &\dots & A_n^\top}^\top$.  We denote the chi-squared distribution with $k$ degrees of freedom by $\chi^2(k)$. Finally, we define the following $N_1T \times n_u$ data matrices:
\begin{align*}
    \Vdata^{(h)} &= \bmat{u_1^{(h)}[0] & \dots & u_1^{(h)}[T-1] & \dots & u_{N_1}^{(h)}[0] & \dots & u_{N_1}^{(h)}[T-1]}^\top, \\
    \Zdata^{(h)} &= \bmat{z_1^{(h)}[0] & \dots & z_1^{(h)}[T-1] & \dots & z_{N_1}^{(h)}[0] & \dots & z_{N_1}^{(h)}[T-1]}^\top.
\end{align*}

\section{Bounding the covariance concentration}\label{appendix: covariance concentration}

We introduce the following version of the Hanson-Wright inequality for sub-gaussian quadratic forms \citep[Theorem 6.3.2]{vershynin2018high}, specialized to the Gaussian case via \cite{laurent2000adaptive}.
\begin{proposition}[Concentration of Gaussian Quadratic Forms]
\label{prop: Hanson-Wright for Gaussians}
Let $z \sim \calN(0, I_{n})$ be a random Gaussian vector, and $R \in \R^{m \times n}$ is an arbitrary fixed matrix. Then the following concentration inequalities hold for all $\varepsilon > 0$:
\begin{align}
    &\prob\brac{\norm{Rz}^2 \geq (1+\varepsilon) \norm{R}_F^2 } \leq \exp\paren{- \frac{1}{4} \min\curly{\frac{\varepsilon^2}{4}, \varepsilon } \frac{\norm{R}_F^2}{\norm{R}^2} } \\
    &\prob\brac{\abs{\norm{Rz}^2  - \norm{R}_F^2}\geq \varepsilon \norm{R}_F^2 } \leq 2\exp\paren{- \frac{1}{4} \min\curly{\frac{\varepsilon^2}{4}, \varepsilon } \frac{\norm{R}_F^2}{\norm{R}^2} },
\end{align}

\end{proposition}

\begin{proof}
    By the rotational invariance of Gaussian random vectors, we have $z^\top R^\top R z \overset{d}{=} \sum_{i=1}^{\min\curly{d,n}} \sigma_i^2 z_i^2$, where $\sigma_i$ is the $i$th singular value of $R$. Noting that $z_i^2$ are i.i.d.\ $\chi^2$ random variables with $1$ degree of freedom, \citet[Lemma 1]{laurent2000adaptive} provides the following upper and lower tail bounds on Gaussian quadratic forms:
    \begin{align*}
        &\prob\brac{\norm{Rz}^2 \geq \norm{R}_F^2 + 2\norm{R^\top R}_F \sqrt{t} + 2 \norm{R}^2 t} \leq \exp(-t) \\
        &\prob\brac{\norm{Rz}^2 \leq \norm{R}_F^2 - 2\norm{R^\top R}_F\sqrt{t} } \leq \exp(-t).
    \end{align*}
    We want to leverage these bounds to prove multiplicative concentration bounds. Focusing on the upper tail first, we observe that
    \begin{align*}
        \norm{R}_F^2 + 2\norm{R^\top R}_F \sqrt{t} + 2 \norm{R}^2 t &\leq \norm{R}_F^2 + 2\norm{R} \norm{R}_F \sqrt{t} + 2 \norm{R}^2 t \\
        &\leq  \norm{R}_F^2 + 4 \max\curly{\norm{R} \norm{R}_F \sqrt{t}, \norm{R}^2 t} \\
        &=: (1 + \varepsilon)\norm{R}_F^2,
    \end{align*}
    such that by construction
    \[
    \prob\brac{\norm{Rz}^2 \geq (1 + \varepsilon) \norm{R}_F^2 } \leq \exp(-t).
    \]
    Solving for the two cases of $t$ in terms of $\varepsilon$ and taking their minimum, we find that
    \[
    \prob\brac{\norm{Rz}^2 \geq (1 + \varepsilon) \norm{R}_F^2 } \leq \exp\paren{- \frac{1}{4} \min\curly{\frac{\varepsilon^2}{4}, \varepsilon } \frac{\norm{R}_F^2}{\norm{R}^2} },
    \]
    which completes the upper tail bound. Repeating this for the lower tail bound, we set $t = \frac{1}{4}\frac{\norm{R}_F^2}{\norm{R}^2} \varepsilon^2$ to get
    \[
    \prob\brac{\norm{Rz}^2 \leq \norm{R}_F^2 - \varepsilon \norm{R}_F^2} \leq \exp\paren{-\frac{1}{4}\frac{\norm{R}_F^2}{\norm{R}^2} \varepsilon^2},
    \]
    which for a given $\varepsilon > 0$ is always upper bounded by the upper tail bound.
    
\end{proof}

The following $\varepsilon$-net argument is standard, and a proof may be found in Chapter 4 of \cite{vershynin2018high}. 
\begin{lemma}
    \label{lem: symmetric covering}
    Let $W$ be and $d\times d $ random matrix, and $\varepsilon \in [0, 1/2)$. Furthermore, let $\calN$ be an $\varepsilon$-net of $\calS^{d-1}$ 
    with minimal cardinality. Then for all $\rho > 0$,
    \[
        \P\brac{\norm{W} > \rho} \leq \paren{\frac{2}{\varepsilon} + 1}^d \max_{x \in \calN} \P\brac{\abs{x^\top W x} > (1-2\epsilon)\rho}.
    \]
\end{lemma}

We now prove a generalized result on the concentration of covariances from \cite{jedra2020finite} in which the covariates are pre and post multiplied by a matrix with orthonormal rows and its transpose, respectively. This demonstrates error bounds scaling with the lower dimension of the orthonormal matrix. 
\begin{restatable}[Modified Lemma 2 of \cite{jedra2020finite}]{lemma}{CovarianceConcentration} \label{lem: covariance concentration} 
    Let $A \in \R^{n \times n}$ satisfy $\rho(A) < 1$
    and let $\Sigma_w \succ 0$ with dimension $n\times n$. Let $\Sigma_x$ solve the discrete Lyapunov equation:
    \[
        \Sigma_x = A \Sigma_x A^\top + \Sigma_w.
    \] Consider drawing $N_1$ trajectories of length $T$ from  a linear system $x_i[t+1] = Ax_i[t] + w_i[t]$, where $w_i[t] \overset{i.i.d}{\sim} \calN(0, \Sigma_w)$ for $i = 1, \dots N_1$ and $t=0, \dots T-1$, and $x_i[0] \overset{i.i.d.}{\sim} \calN(0, \Sigma_x)$ for $i=1,\dots, N_1$.  Let $\Xdata$ be the matrix of stacked state data as in Equation \eqref{eq:stacked xs}. 
    Let $U \in \calO_{n,d}$ with $d \leq n$ be independent of $\Xdata$. Define $M = \paren{N_1 T U^\top \Sigma_x U}^{-1/2}$. Then
    \[
        \norm{(\Xdata U M)^\top \Xdata U M - I} > \max\curly{\varepsilon, \varepsilon^2}
    \]
    holds with probability at most
    \[
        2\exp\paren{-c_1 \varepsilon^2 \frac{N_1 T}{\kappa(\Sigma_x) \calJ(A)^2} + c_2 d}
    \]
    for some absolute constants $c_1$ and $c_2$. 
\end{restatable}
\begin{proof}

Note that we can write 
\[
    \Ex\brac{(\Xdata U)^\top \Xdata U} = \Ex \brac{U^\top \sum_{i=1}^{N_1} \sum_{t=0}^{T-1} x_i[t]x_i[t]^\top U} = N_1 T U^\top \Sigma_x U.
\]
Then
\begin{align*}
    \norm{(\Xdata U M)^\top (\Xdata U M) - I} &= \sup_{v \in \calS^{d-1}} \abs{v^\top \paren{(\Xdata U M)^\top \Xdata U M - I}v} \\
    &= \sup_{v \in \calS^{d-1}} \abs{v^\top \paren{(\Xdata U M)^\top \Xdata U M}v - \Ex v^\top \paren{(\Xdata U M)^\top \Xdata U M}v} \\
    &= \sup_{v \in \calS^{d-1}} \abs{\norm{\Xdata U Mv}^2 - \Ex \norm{\Xdata U Mv}^2} \\
    &\overset{(i)}{=} \sup_{v \in \calS^{d-1}} \abs{\norm{\sigma_{Mv}^\top \tilde \Gamma \xi}^2 - \Ex \norm{\sigma_{Mv}^\top \tilde \Gamma \xi}^2} \\
    &= \sup_{v \in \calS^{d-1}} \abs{\norm{\sigma_{Mv}^\top \tilde \Gamma \xi}^2 -  \norm{\sigma_{Mv}^\top \tilde \Gamma}_F^2},
\end{align*}
where $(i)$ follows by defining
\[
    \Gamma = I_{N_1} \otimes \bmat{ \Sigma_x^{1/2} \\ A \Sigma_x^{1/2} & \Sigma_w^{1/2} && 0 \\ && \ddots \\ A^{T-1}\Sigma_x^{1/2} & \hdots & A \Sigma_w^{1/2} & \Sigma_w^{1/2}  },
\]
\[
    \xi = \vecop \paren{\bmat{\eta_0[-1] & \dots & \eta_0[T-1] & \dots & \eta_{N_1}[0] & \dots & \eta_{N_1}[T-2]}}, 
\] 
where $\eta_i[t] \overset{i.i.d.}{\sim} \calN(0, I)$ for $t \in [-1, T-2]$, and
\[
    \sigma_{Mv} := I_{N_1 T} \otimes (Mv), \qquad \tilde \Gamma := (I_{N_1 T } \otimes U^\top) \Gamma, 
\]
and recalling that $\vecop(U^\top \Xdata^\top) = (I_{N_1 T} \otimes U^\top) \Gamma \xi = \tilde \Gamma \xi$. 
Next oberserve that for any $v \in \calS^{d-1}$, we can apply \Cref{prop: Hanson-Wright for Gaussians} to show that 
\[
    \P\brac{\abs{\norm{\sigma_{Mv}^\top \tilde \Gamma \xi}^2 - \norm{\sigma_{Mv} \tilde \Gamma }_F^2} > \rho \norm{\sigma_{Mv}^\top \tilde \Gamma}_F^2} \leq 2 \exp \paren{-C \min\curly{\rho^2, \rho} \frac{\norm{\sigma_{Mv}^\top \tilde \Gamma}_F^2}{\norm{\sigma_{Mv}^\top \tilde \Gamma}^2}}
\]
for some positive universal constant $C$. 
Next observe that $\norm{\sigma_{Mv} \tilde \Gamma}_F^2 = 1$, and  $\norm{\sigma_{Mv}^\top \tilde \Gamma} \leq \norm{M} \norm{\tilde \Gamma}$, and thus the right hand side reduces to 
\[
    2 \exp \paren{-\frac{C \min\curly{\rho^2, \rho}}{\norm{M}^2\norm{\tilde \Gamma}^2}}.
\]
Applying Lemma~\ref{lem: symmetric covering} with $\varepsilon=\frac{1}{4}$, we have that \[
    \P\brac{\norm{(\Xdata UM)^\top\Xdata UM - I} > \rho} \leq 2\cdot 9^d \exp \paren{-\frac{C \min\curly{ \rho^2/4, \rho/2}}{\norm{M}^2\norm{\tilde \Gamma}^2}}.
\]
Setting $\rho = \max\curly{\varepsilon, \varepsilon^2}$ and rearranging constants provides
\[
    \P\brac{\norm{(\Xdata UM)^\top\Xdata UM - I} > \max\curly{\varepsilon, \varepsilon^2}} \leq 2 \exp \paren{-\frac{c_1\varepsilon^2}{\norm{M}^2\norm{\Gamma}^2} + c_2 d}.
\]
Lastly, note that 
\begin{align*}
    \norm{M} = \frac{\norm{(U^\top \Sigma_x U)^{-1/2}}}{\sqrt{N_1 T}} = \frac{1}{\sqrt{N_1 T}} \frac{1}{ \lambda_d(U^\top \Sigma_x U)^{1/2}} \leq \frac{1}{\sqrt{N_1 T}} \frac{1}{ \lambda_{\min}(\Sigma_x)^{1/2}}
\end{align*}
and
\begin{align*}
   \norm{\tilde \Gamma} &\leq \norm{\Gamma} \\
   &= \norm{\bmat{ \Sigma_x^{1/2} \\ A \Sigma_x^{1/2} & \Sigma_w^{1/2} && 0 \\ && \ddots \\ A^{T-1}\Sigma_x^{1/2} & \hdots & A \Sigma_w^{1/2} & \Sigma_w^{1/2} }} \\ 
   &\leq \norm{\bmat{\Sigma_x^{1/2} \\ \vdots \\ A^{T-1} \Sigma_x^{1/2}}} + \norm{\bmat{\Sigma_w^{1/2} && 0 \\ && \ddots \\ A^{T-2}\Sigma_w^{1/2} & \hdots & A \Sigma_w^{1/2} & \Sigma_w^{1/2}}} \\ 
   &\leq \sum_{s=0}^{T-1} \norm{A^s} \norm{\Sigma_x}^{1/2} + \norm{\bmat{I && 0 \\ && \ddots \\ A^{T-2} & \hdots & A  & I}} \norm{\Sigma_w}^{1/2} \\
   & \leq 2 \sum_{s \geq 0} \norm{A^s} \norm{\Sigma_x}^{1/2} \\
   &= 2 \calJ(A) \norm{\Sigma_x}^{1/2}. 
\end{align*}
where the final inequality follows by applying Lemma 5 of \cite{jedra2020finite} and the fact that $\Sigma_w \preceq \Sigma_x$. The lemma then follows from the fact that $\kappa(\Sigma_x) = \frac{\norm{\Sigma_x}}{\lambda_{\min}(\Sigma_x)}$.
\end{proof}

The previous lemma can now be applied to show concentration of the empirical covariance for both the source and target task. 

\begin{restatable}{lemma}{SourceCovarianceConcentration}
    \label{lem: source covariance concentration}
    Suppose $N_1 T \geq C \max_h \calJ\paren{A + B K^{(h)}}^2 \kappa\paren{\Sigma_x^{(h)}} (n_x + \log(H/\delta))$ for $\delta \in (0,1)$, where $C$ is a universal numerical constant. Then with probability at least $1-\frac{\delta}{10}$ over the states $\Xdata^{(1)}, \dots, \Xdata^{(H)}$ in the source tasks, we have
    \[
        0.9 \Sigma_x^{(h)} \preceq \frac{1}{N_1 T} \paren{\Xdata^{(h)}}^\top \Xdata^{(h)} \preceq 1.1 \Sigma_x^{(h)}, \qquad \forall h \in [H].
    \]
\end{restatable}

\begin{proof}
Applying Lemma~\ref{lem: covariance concentration} with $U = I$ and $\varepsilon = 0.1$,
as long as $C \geq 100\frac{\max\curly{c_2, \log(1/20)}}{c_1}$,
then
for any task $h \in [H]$, we have that
$$\norm{(\Xdata^{(h)} M)^\top \Xdata^{(h)} M - I} \leq 0.1,$$ 
with probability at least
$1-2\exp\paren{-\frac{0.01 c_1 N_1 T}{\calJ(A + B K^{(h)})^2 \kappa(\Sigma_x^{(h)})} + c_2 n_x} \geq 1-\frac{\delta}{10 H}$.
This may be written equivalently as 
\[
    0.9 \Sigma_x^{(h)} = 0.9 \frac{M^{-2}}{N_1 T} \preceq \frac{\Xdata^{(h)\top} \Xdata^{(h)}}{N_1 T} \preceq 1.1 \frac{M^{-2}}{N_1 T} = 1.1 \Sigma_x^{(h)}. 
\]
Taking a union bound over the $H$ tasks provides the desired result.  
\end{proof}
\begin{restatable}{lemma}{TargetCovarianceConcentration}
    \label{lem: target covariance concentration}
    Suppose $N_2 T \geq  C\calJ\paren{A + B K^{(H+1)}}^2 \kappa\paren{\Sigma_x^{(H+1)}}(k + \log(1/\delta))$ for $\delta \in (0,1)$, where $C$ is a universal numerical constant. Then for any given matrix $\Phi \in \R^{n_x \times 2k}$ independent of $\Xdata^{(H+1)}$, with probability at least $1-\frac{\delta}{10}$ over $\Xdata^{(H+1)}$ we have
    \[
        0.9 \Phi^\top \Sigma_x^{(H+1)} \Phi \preceq \frac{1}{N_2 T} \Phi^\top \paren{\Xdata^{(H+1)}}^\top \Xdata^{(H+1)} \Phi \preceq 1.1 \Phi^\top \Sigma_x^{(H+1)} \Phi. 
    \]
\end{restatable}
\begin{proof}
Let $\Phi = U S V^\top$ be the singular value decomposition of $\Phi$ with $U \in \R^{n_x \times 2k}$. Applying Lemma~\ref{lem: covariance concentration} with $\varepsilon = 0.1$, we have that 
\begin{align}
    \label{eq: target concentration}
    0.9 U^\top \Sigma_x^{(H+1)} U = 0.9 \frac{M^{-2}}{N_2 T} \preceq \frac{U^\top \Xdata^{(H+1)\top} \Xdata^{(H+1)} U}{N_2 T} \preceq 1.1 \frac{M^{-2}}{N_2 T} = 1.1 U^\top \Sigma_x^{(H+1)} U
\end{align}
with probability at least $1-2\exp\paren{-0.01 c_1 \frac{N_2 T}{\calJ(A + B K^{(H+1)})^2 \kappa(\Sigma_x^{(H+1)})} + 2 c_2 k} \geq 1 - \frac{\delta}{10}$ as long as $C \geq 100\frac{\max\curly{c_2, \log(1/20)}}{c_1}$. Left and right multiplying Equation \eqref{eq: target concentration} by $V^\top S$ and $SV$ respectively results in 
\[
    0.9 \Phi^\top \Sigma_x^{(H+1)} \Phi \preceq \frac{\Phi^\top \Xdata^{(H+1)\top} \Xdata^{(H+1)} \Phi}{N_2 T} \preceq 1.1 \Phi^\top \Sigma_x^{(H+1)} \Phi
\]
with probability at least $1-\frac{\delta}{10}$.
\end{proof}

\section{Data Guarantees}\label{appendix: data guarantees}
We will now use the covariance concentration results to show concentration of the source and target controllers to the optimal controllers. We begin by recalling Lemma A.5 of \cite{du2020few}.
\begin{lemma}
    \label{lem: du A.5}
    There exists a subset $\calN \subset \calO_{d_1, d_2}$ that is an $\varepsilon$-net of $\calO_{d_1,d_2}$ in Frobenius norm such that $\abs{\calN} \leq \paren{\frac{6\sqrt{d_2}}{\varepsilon}}^{d_1 d_2}$.
\end{lemma}
Also, note the following fact.
\begin{fact}
\label{fact: self normalized LS opt}
For size conforming $Z, X$, with $X$ full column rank,
\begin{align*}
    \sup_{F} \left[4 \langle Z, XF^\top \rangle - \norm{XF^\top}_F^2\right] = 4 \norm{(X^\top X)^{-1/2} X^\top Z}_F^2 = 4 \norm{P_{X} Z}_F^2.
\end{align*}
\end{fact}
The following result on perturbation of projection matrices may be found in \cite{chen2016perturbation, xu2020perturbation}.
\begin{lemma}[Perturbation of projection matrices]
\label{lemma:projection_perturbation}
Let $A, B$ be size conforming matrices both of the same rank. We have:
$$
    \norm{P_A-P_B} \leq \min\left\{ \norm{A^\dag}, \norm{B^\dag} \right\} \norm{A-B}.
$$
\end{lemma}


\begin{lemma}
\label{lemma:yasin_supermartingale}
Let $(x_t)_{t \geq 1}$ be a $\R^d$-valued process adapted to a filtration $(\calF_t)_{t \geq 1}$.
Let $(\eta_t)_{t \geq 1}$ be a $\R^m$-valued process adapted to $(\calF_t)_{t \geq 2}$.
Suppose that $(\eta_t)_{t \geq 1}$ is a $\sigma$-sub-Gaussian
martingale difference sequence, i.e.,:
\begin{align}
    \Ex[ \eta_t \mid \calF_t ] &= 0, \\
    \Ex[ \exp(\lambda \langle v, \eta_t \rangle ) \mid \calF_t ] &\leq \exp\paren{\frac{\lambda^2 \sigma^2 \norm{v}^2}{2}} \:\:\forall \calF_t \text{-measurable } \lambda \in \R, v \in \R^m.
\end{align}
For $\Lambda \in \R^{m \times d}$,
let $(M_t(\Lambda))_{t \geq 1}$ be the $\R$-valued process:
\begin{align}
    M_t(\Lambda) = \exp\left( \frac{1}{\sigma}\sum_{i=1}^{t} \langle \Lambda x_i, \eta_i \rangle - \frac{1}{2}\sum_{i=1}^{t} \norm{\Lambda x_i}^2   \right).
\end{align}
The process $(M_t(\Lambda))_{t \geq 1}$
satisfies $\Ex[ M_t(\Lambda) ] \leq 1$ for all $t \geq 1$.
\end{lemma}
\begin{proof}
Let $M_0(\Lambda) := 1$.
Fix $t \geq 1$. Observe:
\begin{align*}
    \Ex[ M_t(\Lambda) ] &= \Ex[ \Ex[M_t(\Lambda) \mid \calF_t]] = \Ex\left[  M_{t-1}(\Lambda) \Ex\left[\exp\paren{ \frac{1}{\sigma} \langle \Lambda x_t, \eta_t \rangle } \,\bigg|\, \calF_t\right] \exp\paren{-\frac{1}{2}\norm{\Lambda x_t}^2} \right] \\
    &\leq \Ex[ M_{t-1}(\Lambda) ].
\end{align*}
\end{proof}

The following result generalizes the self-normalized martingale inequality from \cite{abbasi2011online} to handle multiple matrix valued self-normalized martingales. 
\begin{proposition}[Generalized self-normalized martingale inequality]
\label{prop:yasin_multiprocesses}
Fix $H \in \N_+$. For $h \in [H]$,
let $(x_t^h, \eta_t^h)_{t \geq 1}$ be a $\R^d \times \R^m$-valued process
and $(\calF_t^h)_{t \geq 1}$ be a filtration such that
$(x_t^h)_{t \geq 1}$ is adapted to $(\calF_t^h)_{t \geq 1}$,
$(\eta_t^h)_{t \geq 1}$ is adapted to $(\calF_t^h)_{t \geq 2}$, and 
$(\eta_t^h)_{t \geq 1}$ is a $\sigma$-sub-Gaussian martingale difference sequence.
Suppose that for all $h_1 \neq h_2$, the process
$(x_t^{h_1}, \eta_t^{h_1})$ is independent
of $(x_t^{h_2}, \eta_t^{h_2})$.
Fix (non-random) positive definite matrices $\{V^h\}_{h=1}^{H}$.
For $t \geq 1$ and $h \in [H]$, define:
\begin{align}
    \bar{V}_t^h := V^h + V^h_t, \:\: V^h_t := \sum_{i=1}^{t} x_i x_i^\top, \:\: S_t^h := \sum_{i=1}^{t} x_i \eta_i^\top.
\end{align}
For any fixed $T \in \N_+$,
with probability at least $1-\delta$:
\begin{align}
    \sum_{h=1}^{H} \norm{ (\bar{V}^h_T)^{-1/2} S^h_T }_F^2 \leq 2 \sigma^2 \left[\sum_{h=1}^{H} \log\mathrm{det}( (\bar{V}_T^h)^{m/2} (V^h)^{-m/2}) + \log(1/\delta)\right].
\end{align}
\end{proposition}
\begin{proof}
By rescaling, we assume that $\sigma=1$ without loss
of generality.
For $h \in [H]$, define:
\begin{align}
    M_T^h(\Lambda) := \exp\left( \sum_{t=1}^{T} \langle \Lambda x_t^h, \eta_t^h \rangle - \frac{1}{2}\sum_{t=1}^{T} \norm{\Lambda x_t^h}^2   \right), \:\: \Lambda \in \R^{m \times d}.
\end{align}
By \Cref{lemma:yasin_supermartingale},
$\Ex[ M_T^h(\Lambda) ] \leq 1$ for any fixed $\Lambda$.
Now, we use the method of mixtures argument from
\cite{abbasi2011online}.
Let $\Gamma \in \R^{m \times d}$ be a random matrix with
i.i.d.\ $N(0, 1)$ entries, independent of everything else.
By the tower property, $\Ex[ M_T^h(\Gamma) ] = \Ex[ \Ex[M_T^h(\Gamma)] \mid \Gamma] \leq 1$.
On the other hand, let us compute $\Ex[ M_T^h(\Gamma) \mid \calF^h_\infty]$. Let $p(\Lambda)$ denote the
density of $\Gamma$,
and also let $p(g)$ denote the density of a $d$-dimensional isotropic normal vector. Let us momentarily
drop the superscript $h$ since the computation
is identical for every $h$. 
First, we note that the proof of Theorem 3 in
\cite{abbasi2011online} shows the following identity
for any fixed $u \in \R^d$, $V \in \R^{m \times m}$ positive semidefinite, and $V_0 \in \R^{m \times m}$ positive definite:
\begin{align*}
    \int \exp\paren{ \langle g, u \rangle - \frac{1}{2} \norm{g}^2_V} p(g) \,\rmd g = \paren{ \frac{\mathrm{det}(V_0)}{\mathrm{det}(V_0 + V)} }^{1/2} \exp\paren{\frac{1}{2}\norm{u}^2_{(V+V_0)^{-1}}}.
\end{align*}
With this identity and independence of the
rows of $\Gamma$:
\begin{align*}
    \Ex[ M_T(\Gamma) \mid \calF_\infty] &= \int \exp\paren{ \langle \Lambda^\top, S_T \rangle - \frac{1}{2} \norm{ \Lambda^\top }_{V_t}^2 } p(\Lambda) \,\rmd\Lambda \\
    &= \int \exp\paren{ \sum_{i=1}^{m} \langle \Lambda^\top e_i, S_T e_i \rangle - \frac{1}{2} \norm{\Lambda^\top e_i}^2_{V_t} } p(\Lambda) \,\rmd\Lambda \\
    &= \prod_{i=1}^{m} \int \exp\paren{ \langle g, S_T e_i \rangle - \frac{1}{2} \norm{g}^2_{V_t} } p(g) \,\rmd g \\
    &= \paren{ \frac{\mathrm{det}(V)}{\mathrm{det}(V + V_T)} }^{m/2} \exp\paren{ \frac{1}{2} \sum_{i=1}^{m} \norm{ S_T e_i }^2_{(V + V_T)^{-1}}  } \\
    &= \paren{ \frac{\mathrm{det}(V)}{\mathrm{det}(V + V_T)} }^{m/2} \exp\paren{ \frac{1}{2}\norm{(V+V_T)^{-1/2} S_T}_F^2 }.
\end{align*}
That is, for every $h \in [H]$, we have:
\begin{align*}
    \Ex\left[ \paren{ \frac{\mathrm{det}(V^h)}{\mathrm{det}(\bar{V}^h_T)} }^{m/2} \exp\paren{ \frac{1}{2}\norm{(\bar{V}_T^h)^{-1/2} S_T}_F^2} \right] \leq 1.
\end{align*}
Now by Markov's inequality and
independence of the processes across $h$:
\begin{align*}
    &\P\left( \sum_{h=1}^{H}\norm{(\bar{V}_T^h)^{-1/2} S_T}_F^2  > 2 \left[\sum_{h=1}^{H} \log\mathrm{det}( (\bar{V}_T^h)^{m/2} (V^h)^{-m/2}) + \log(1/\delta)\right] \right)\\
    &= \P\left( \exp\paren{ \frac{1}{2}\sum_{h=1}^{H}\norm{(\bar{V}_T^h)^{-1/2} S_T}_F^2  } > \delta^{-1} \prod_{h=1}^{H} \paren{ \frac{\mathrm{det}(\bar{V}_T^h)}{\mathrm{det}(V^h)}  }^{m/2} \right) \\
    &= \P\left(\prod_{h=1}^{H} \paren{ \frac{\mathrm{det}(V^h)}{\mathrm{det}(\bar{V}_T^h)}  }^{m/2} \exp\paren{ \frac{1}{2}\norm{(\bar{V}_T^h)^{-1/2} S_T}_F^2  } > \delta^{-1}  \right) \\
    &\leq \delta \Ex\left[\prod_{h=1}^{H} \paren{ \frac{\mathrm{det}(V^h)}{\mathrm{det}(\bar{V}_T^h)}  }^{m/2} \exp\paren{ \frac{1}{2}\norm{(\bar{V}_T^h)^{-1/2} S_T}_F^2  }  \right]  \\
    &= \delta \prod_{h=1}^{H} \Ex\left[ \paren{ \frac{\mathrm{det}(V^h)}{\mathrm{det}(\bar{V}_T^h)}  }^{m/2} \exp\paren{ \frac{1}{2}\norm{(\bar{V}_T^h)^{-1/2} S_T}_F^2 } \right] \\
    &\leq \delta.
\end{align*}
\end{proof}

\begin{remark}
The original self-normalized martingale bound from \citet{abbasi2011online} 
holds for an arbitrary stopping time, 
and consequently \emph{uniformly} for all time $t \in \N$ by a simple reduction.
In contrast, \Cref{prop:yasin_multiprocesses} holds for a fixed $t \in \N$,
which suffices for our purposes.
\end{remark}

\begin{lemma}[source controller concentration]
    \label{lem: source controller concentration}
    Under the setting and data assumptions of \Cref{thm: target task excess risk bound}, with probability at least $1-\frac{\delta}{5}$,
    \begin{align*}
        \sum_{h=1}^H \norm{\Xdata^{(h)} \paren{\hat F^{(h)} \hat \Phi - F^{(h)}_\star \Phi_\star}^\top}_F^2 \lesssim \sigma_z^2 \paren{k m H + kn \log\paren{N_1 T \frac{\bar \lambda}{\underline{\lambda}}}  + \log\paren{\frac{1}{\delta}}}.
    \end{align*}
\end{lemma}
\begin{proof}
    Before we start the proof, we first define 
    two subsets:
    \begin{align*}
        \Theta_1 &:= \{ (F_1 \Phi, \dots, F_h \Phi) \mid F_i \in \R^{n_u \times k}, \:\: \Phi \in \R^{k \times n_x} \}, \\
        \Theta_2(a, b) &:= \{ (F_1 \Phi, \dots, F_h \Phi) \mid F_i \in \R^{n_u \times b}, \:\: \Phi^\top \in \calO_{a,b} \}, \:\: a \geq b.
    \end{align*}
    It is easy to see that $\Theta_1 = \Theta_2(n_x, k)$.
    Furthermore, a simple argument shows that
    $$\Theta_2(n_x, k) - \Theta_2(n_x, k) \subset \Theta_2(n_x, 2k),$$
    where the minus sign indicates the Minkowski difference
    of two sets.

    Throughout this proof, we will condition on the following event, which holds with probability at least  $1-\frac{\delta}{10}$ by Lemma~\ref{lem: source covariance concentration}:
    \begin{align}
        \label{eq: source covariance concentration}
        0.9 \Sigma_x^{(h)} \preceq \frac{1}{N_1 T} \paren{\Xdata^{(h)}}^\top \Xdata^{(h)} \preceq 1.1 \Sigma_x^{(h)}, \qquad \forall h \in [H].
    \end{align}
    
    By optimality of $\hat F^{(1)}, \hat F^{(2)}, \dots, \hat F^{(H)}$ and $\hat \Phi$ for Problem \eqref{eq: pretraiing}, we have that
    \begin{align*}
        \sum_{h=1}^H \norm{\Vdata^{(h)} - \Xdata^{(h)} (\hat F^{(h)} \hat \Phi)^\top}_F^2 \leq \sum_{h=1}^H  \norm{\Vdata^{(h)} - \Xdata^{(h)} (F^{(h)}_\star  \Phi_\star)^\top}_F^2.
    \end{align*}
    Substituting in $\Vdata^{(h)} = \Xdata^{(h)} (F^{(h)}_\star \Phi_\star)^\top + \Zdata^{(h)}$
    and re-arranging yields the basic inequality:
    \begin{align}
        \label{eq: data weighted source controller gap}
        \sum_{h=1}^H \norm{ \Xdata^{(h)} (\hat F^{(h)} \hat \Phi - F^{(h)}_\star \Phi_\star)^\top}_F^2 \leq 2\sum_{h=1}^H  \trace\paren{\paren{\Zdata^{(h)}}^\top \Xdata^{(h)} (F^{(h)}_\star \Phi_\star - \hat F^{(h)} \hat \Phi )^\top}.
    \end{align}
    Let $\Delta^{(h)} := \hat F^{(h)} \hat \Phi - F^{(h)}_\star \Phi_\star$.
    Multiplying the basic inequality above by two and re-arranging again,
    we obtain the \emph{offset basic inequality} \citep{liang2015learning}:
    \begin{align*}
        \sum_{h=1}^{H} \norm{ \Xdata^{(h)} (\Delta^{(h)})^\top }^2_F &\leq \sum_{h=1}^{H} 4 \trace\paren{ {(\Zdata^{(h)})^\top}  {\Xdata^{(h)}} {(\Delta^{(h)})^\top} } - \norm{ \Xdata^{(h)} (\Delta^h{(h)})^\top }_F^2 \\
        &\leq \sup_{\{\Delta^{(h)}\}_{h=1}^{H} \in \Theta_1 - \Theta_1}\sum_{h=1}^{H} 4 \trace\paren{ {(\Zdata^{(h)})^\top}  {\Xdata^{(h)}} {(\Delta^{(h)})^\top} } - \norm{ \Xdata^{(h)} (\Delta^h{(h)})^\top }_F^2 \\
        &= \sup_{\{\Delta^{(h)}\}_{h=1}^{H} \in \Theta_2(n_x, k) - \Theta_2(n_x, k)}\sum_{h=1}^{H} 4 \trace\paren{ {(\Zdata^{(h)})^\top}  {\Xdata^{(h)}} {(\Delta^{(h)})^\top} } - \norm{ \Xdata^{(h)} (\Delta^h{(h)})^\top }_F^2 \\ 
        &\leq \sup_{\{\Delta^{(h)}\}_{h=1}^{H} \in \Theta_2(n_x, 2k)}  \sum_{h=1}^{H} 4 \trace\paren{ {(\Zdata^{(h)})^\top}  {\Xdata^{(h)}} {(\Delta^{(h)})^\top} } - \norm{ \Xdata^{(h)} (\Delta^h{(h)})^\top }_F^2 \\
        &= \sup_{\Phi^\top \in \calO_{n_x, 2k}} \sum_{h=1}^{H} \sup_{F_h \in \R^{n_u \times 2k}}\left[4 \trace\paren{ {(\Zdata^{(h)})^\top}  {\Xdata^{(h)}} \Phi^\top F_h^\top } - \norm{ \Xdata^{(h)} \Phi^\top F_h^\top }_F^2  \right] \\
        &= 4 \sup_{\Phi^\top \in \calO_{n_x, 2k}} \sum_{h=1}^{H} \norm{(\Phi (\Xdata^{(h)})^\top \Xdata^{(h)} \Phi^\top)^{-1/2} \Phi (\Xdata^{(h)})^\top \Zdata^{(h)}   }_F^2.
    \end{align*}
    The last equality above follows from
    \Cref{fact: self normalized LS opt}.
    We now derive an upper bound on the
    sum for a fixed $\Phi^\top \in \calO_{n_x, 2k}$.
    To do this, we invoke 
    \Cref{prop:yasin_multiprocesses} with the trajectories within each task concatenated in sequence: $V^h \gets 0.9 N_1 T \Phi \Sigma_x^{(h)} \Phi^\top$,
    $x_t^h \gets \Phi x_i^{(h)}[t]$,
    and $\eta_t^h \gets z_i^{(h)}[t]$.
    Note that since 
    $\Phi (\Xdata^{(h)})^\top \Xdata^{(h)} \Phi^\top \succeq V^h$, we then have that:
    $$( \Phi (\Xdata^{(h)})^\top \Xdata^{(h)} \Phi^\top )^{-1} \preceq 2 ( \Phi (\Xdata^{(h)})^\top \Xdata^{(h)} \Phi^\top + V^h )^{-1}.$$
    Consequently, with probability at least $1-\delta',$
    \begin{align}
        \sum_{h=1}^{H} &\norm{(\Phi (\Xdata^{(h)})^\top \Xdata^{(h)} \Phi^\top)^{-1/2} \Phi (\Xdata^{(h)})^\top \Zdata^{(h)}   }_F^2  \nonumber \\
        &\lesssim  \sum_{h=1}^{H} \norm{(\Phi (\Xdata^{(h)})^\top \Xdata^{(h)} \Phi^\top + V^h)^{-1/2} \Phi (\Xdata^{(h)})^\top \Zdata^{(h)}   }_F^2 \nonumber\\
        &\lesssim \sigma^2 \brac{\sum_{h=1}^H \log \mathrm{det} \paren{\paren{\Phi \paren{\Xdata^{(h)}}^\top \Xdata^{(h)} \Phi^\top + V^h}^{n_u/2}\paren{V^h}^{-n_u/2}} + \log(1/\delta')} \nonumber\\
        &=  \sigma^2 \brac{\frac{n_u}{2}\sum_{h=1}^H \log \mathrm{det} \paren{\paren{\Phi \paren{\Xdata^{(h)}}^\top \Xdata^{(h)} \Phi^\top + V^h}\paren{V^h}^{-1}} + \log(1/\delta')} \nonumber\\
        &=  \sigma^2 \brac{\frac{n_u}{2}\sum_{h=1}^H \log \mathrm{det} \paren{\frac{1}{0.9 N_1 T}\paren{\Phi \paren{\Xdata^{(h)}}^\top \Xdata^{(h)} \Phi^\top} \paren{\Phi \Sigma_x^{(h)} \Phi^\top}^{-1} + I_{2k}} + \log(1/\delta')} \nonumber\\
        &\leq  \sigma^2 \brac{\frac{n_u}{2}\sum_{h=1}^H \log \mathrm{det} \paren{\frac{1.1}{0.9}+1} I_{2k} + \log(1/\delta')} \nonumber\\
        &\lesssim \sigma_z^2( kH n_u + \log(1/\delta')). \label{eq:apply self normalized}
    \end{align}
    This holds for a fixed $\Phi^\top \in \calO_{n_x, 2k}$, so it remains to union bound over $\calO_{n_x, 2k}$.
    Let $\{\Phi_i^\top\}_{i=1}^{N}$ be an $\varepsilon$-net of $\calO_{n_x, 2k}$ in the spectral norm of resolution to be determined.
    For $\Phi^\top \in \calO_{n_x, 2k}$, let $\Phi_i^\top$ denote the nearest element in the cover.
    By triangle inequality and $(a+b)^2 \leq 2(a^2+b^2)$:
    \begin{align*}
        \sum_{h=1}^{H} &\norm{(\Phi (\Xdata^{(h)})^\top \Xdata^{(h)} \Phi^\top)^{-1/2} \Phi (\Xdata^{(h)})^\top \Zdata^{(h)}   }_F^2 \\
        &=  \sum_{h=1}^{H} \norm{P_{\Xdata^{(h)} \Phi} \Zdata^{(h)}   }_F^2 \\
        &\leq 2 \sum_{h=1}^{H} \norm{ (P_{\Xdata^{(h)} \Phi} - P_{\Xdata^{(h)} \Phi_i}) \Zdata^{(h)} }_F^2 + 2\sum_{h=1}^{H} \norm{P_{\Xdata^{(h)} \Phi_i} \Zdata^{(h)}}^2_F \\
        &\leq 2 \sum_{h=1}^{H} \norm{P_{\Xdata^{(h)} \Phi} - P_{\Xdata^{(h)} \Phi_i}}^2_2 \norm{\Zdata^{(h)}}_F^2 + 2\sum_{h=1}^{H} \norm{P_{\Xdata^{(h)} \Phi_i} \Zdata^{(h)}}^2_F.
    \end{align*}
    By \Cref{lemma:projection_perturbation},
    \begin{align*}
        \norm{P_{\Xdata^{(h)} \Phi} - P_{\Xdata^{(h)} \Phi_i}} &\leq \norm{( \Phi (\Xdata^{(h)})^\top \Xdata^{(h)} \Phi^\top )^{-1} \Phi (\Xdata^{(h)})^\top  } \norm{\Xdata^{(h)} (\Phi - \Phi_i) } \\
        &\leq \frac{1.1}{0.9} \frac{ \lambda_{\max}(\Sigma_x^{(h)})}{\lambda_{\min}(\Sigma_x^{(h)})} \varepsilon \lesssim \frac{\overline{\lambda}}{\underline{\lambda}} \varepsilon.
    \end{align*}
    Next, we need to bound
    $\sum_{h=1}^{H} \norm{\Zdata^{(h)}}^2_F$.
    Because each $z_i^{(h)}[t]$ is i.i.d.\ $\calN(0, \sigma_z^2 I)$, this sum is distributed as a
    $\sigma_z^2 \psi$, where $\psi$ is a $\chi^2$ random variable with $H N_1 T n_u$ degrees of freedom.
    Hence by standard $\chi^2$ tail bounds, with probability at least $1-\delta/20$,
    $\sum_{h=1}^{H} \norm{\Zdata^{(h)}}_F^2 \lesssim \sigma_z^2 (HN_1 T n_u + \log(1/\delta))$.
    This prompts us to select
    $\varepsilon \lesssim \frac{k}{N_1 T \overline{\lambda}/\underline{\lambda}}$,
    from which we conclude:
    $$
        \sum_{h=1}^{H} \norm{P_{\Xdata^{(h)} \Phi} - P_{\Xdata^{(h)} \Phi_i}}^2\norm{\Zdata^{(h)}}_F^2 \lesssim \sigma_z^2 k H n_u.
    $$
    By \Cref{lem: du A.5}, we can then bound the
    size of the $\varepsilon$-covering of $\calO_{n_x, 2k}$ by:
    $$
        N \leq \paren{ \frac{c \sqrt{k}}{k} N_1 T \frac{\overline{\lambda}}{\underline{\lambda}}  }^{2n_x k}.
    $$
    So now we take $\delta' = \delta/(20N)$, and conclude. In particular, by union bounding over the elements of $\curly{\Phi_i}_{i=1}^N$, we have that with probability at least $1-\frac{\delta}{5}$,
    \begin{align*}
        \sup_{\Phi^\top \in \calO_{n_x, 2k}} & \sum_{h=1}^{H} \norm{(\Phi (\Xdata^{(h)})^\top \Xdata^{(h)} \Phi^\top)^{-1/2} \Phi (\Xdata^{(h)})^\top \Zdata^{(h)}   }_F^2 \\
        &\lesssim \sigma_z^2 \paren{ k H n_u  + n_x k \log\paren{N_1 T \frac{\bar \lambda}{\underline \lambda} } + \log \paren{\frac{1}{\delta}}}.
    \end{align*}
    The probability of $1-\frac{\delta}{5}$ arises from union bounding over Equation \eqref{eq: source covariance concentration}, which holds with probability at least $1-\frac{\delta}{10}$, the bound on $\sum_{h=1}^H \norm{\Zdata^{(h)}}_F^2$ which holds with probability at least $1-\frac{\delta}{20}$, and the bound in Equation \eqref{eq:apply self normalized}, which holds for all $\Phi_i \in \curly{\Phi_i}_{i=1}^N$ with probability at least $1-\frac{\delta}{20}$.

\end{proof}

\begin{lemma}[Lemma A.7 in \cite{du2020few}]
    \label{lem: du A.7}
    For any two matrices $A_1$ and $A_2$ with the same number of columns which satisfy $A_1 ^\top A_1 \succeq A_2^\top A_2$, then for any matrix $B$, we have 
    \[
        A_1^\top P_{A_1 B}^\perp A_1 \succeq A_2^\top P_{A_2 B}^\perp A_2.
    \]
    As a consequence, for any matrices $B$ and $B'$, we have
    \[
        \norm{P_{A_1 B}^\perp A_1 B'}_F^2 \geq \norm{P_{A_2 B}^\perp A_2 B'}_F^2.
    \]
\end{lemma}

\begin{lemma}[target controller concentration]
    \label{lem: target controller concentration}
    Under the setting and data assumptions of Theorem~\ref{thm: target task excess risk bound}, with probability at least $1-\frac{2 \delta}{5}$, 
    \begin{align*}
        \frac{1}{N_2T}  &\norm{\Xdata^{(H+1)} \paren{\hat F^{(H+1)} \hat\Phi - F^{(H+1)}_\star \Phi_\star}^\top}_F^2 \lesssim  \bar \sigma_z^2 \paren{\frac{k n_x \log\paren{N_1 T \frac{\bar \lambda}{\underline{\lambda}}} }{ cN_1 T H} + \frac{kn_u +\log(\frac{1}{\delta})}{N_2 T}}. 
    \end{align*}
\end{lemma}
\begin{proof}
By the optimality of $\hat \Phi, \hat F^{(1)}, \dots, \hat F^{(H)}$ for Problem \eqref{eq: pretraiing},  we know that 
\[\hat F^{(h)} = \paren{\hat \Phi \paren{\Xdata^{(h)}}^\top \Xdata^{(h)} \hat \Phi^\top}^{-1} \hat \Phi \paren{\Xdata^{(h)}}^\top \Vdata^{(h)}. \]
Therefore, $\Xdata^{(h)} \paren{\hat F^{(h)} \hat \Phi}^\top  = P_{\Xdata^{(h)} \hat \Phi^\top} \Vdata^{(h)}  = P_{\Xdata^{(h)} \hat \Phi^\top} \paren{\Xdata^{(h)} \paren{F^{(h)}_\star \Phi_\star}^\top + \Zdata^{(h)}}$. Then by applying Lemma~\ref{lem: source controller concentration}, we have that with probability at least $1-\frac{\delta}{5}$,
\begin{align*}
     \bar \sigma_z^2 &\paren{k m H + kn \log\paren{N_1 T \frac{\bar \lambda}{\underline{\lambda}}}  + \log\paren{\frac{1}{\delta}}} \gtrsim \sum_{h=1}^H \norm{\Xdata^{(h)} \paren{\hat F^{(h)} \hat \Phi - F^{(h)}_\star \Phi_\star}^\top}_F^2 \nonumber\\
     & = \sum_{h=1}^H \norm{P_{\Xdata^{(h)} \hat \Phi^\top} \paren{\Xdata^{(h)} \paren{F^{(h)}_\star \Phi_\star}^\top + Z^{(h)}} - \Xdata^{(h)} \paren{F^{(h)}_\star \Phi_\star}^\top}_F^2 \nonumber\\
     &= \sum_{h=1}^H \norm{(P_{\Xdata^{(h)} \hat \Phi^\top} - I) \Xdata^{(h)} \paren{F^{(h)}_\star \Phi_\star}^\top + P_{\Xdata^{(h)} \hat \Phi^\top} \Zdata^{(h)}}_F^2 \nonumber\\
     &= \sum_{h=1}^H \norm{P_{\Xdata^{(h)} \hat \Phi^\top}^\perp \Xdata^{(h)} \paren{F^{(h)}_\star \Phi_\star}^\top}_F^2 + \norm{P_{\Xdata^{(h)} \hat \Phi^\top} \Zdata^{(h)}}_F^2 \nonumber\\
      &\geq  \sum_{h=1}^H \norm{P_{\Xdata^{(h)} \hat \Phi^\top}^\perp \Xdata^{(h)} \paren{F^{(h)}_\star \Phi_\star}^\top}_F^2 \nonumber\\
      &\overset{(i)}{\geq} 0.9 N_1T \sum_{h=1}^H \norm{P_{\paren{\Sigma^{(h)}}^{1/2} \hat \Phi^\top}^\perp \paren{\Sigma^{(h)}}^{1/2} \paren{F^{(h)}_\star \Phi_\star}^\top}_F^2 \nonumber \\
      &\overset{(ii)}{\geq} 0.9cN_1T\sum_{h=1}^H \norm{P_{\paren{\Sigma^{(H+1)}}^{1/2} \hat \Phi^\top}^\perp \paren{\Sigma^{(H+1)}}^{1/2} \paren{F^{(h)}_\star \Phi_\star}^\top}_F^2 \nonumber \\
\end{align*}
where $(i)$ follow from the inequality in Lemma~\ref{lem: source covariance concentration} in addition to Lemma~\ref{lem: du A.7}, while $(ii)$ follows from the inequality in \eqref{eq: target task covariance coverage} (note that we have already conditioned on this event through Lemma~\ref{lem: source controller concentration}) in addition to Lemma~\ref{lem: du A.7}. If we define
\begin{align*}
    \Fdata_\star = \bmat{F_\star^{(1)} \\ \vdots \\ F_\star^{(H)}},
\end{align*}
we can write the above result concisely as
\begin{align}
     \bar \sigma_z^2 &\paren{k m H + kn \log\paren{N_1 T \frac{\bar \lambda}{\underline{\lambda}}}  + \log\paren{\frac{1}{\delta}}} \gtrsim 0.9cN_1 T\norm{P_{\paren{\Sigma^{(H+1)}}^{1/2} \hat \Phi^\top}^\perp \paren{\Sigma^{(H+1)}}^{1/2} \Phi_\star^\top \Fdata_\star^\top}_F^2 \label{eq: target data projection bound}
\end{align}

Now, letting $\Phi = \bmat{ \hat \Phi^\top & \hat \Phi_\star^\top}$, Lemma~\ref{lem: target covariance concentration} gives us $\frac{1}{N_2 T} \Phi^\top \paren{\Xdata^{(H+1)}}^\top \Xdata^{(H+1)} \Phi \preceq 1.1 \Phi^\top \Sigma_x^{(H+1)} \Phi$. Combining with Lemma~\ref{lem: du A.7}, we have that with probability at least $1-\frac{\delta}{10}$, 
\begin{align*}
    1.1 \norm{P_{\paren{\Sigma^{(H+1)}}^{1/2} \hat \Phi^\top }^\perp \paren{\Sigma^{(H+1)}}^{1/2} \Phi_\star^\top \Fdata_\star^\top}_F^2 \geq \frac{1}{N_2 T}  \norm{P_{\Xdata^{(H+1)} \hat \Phi^\top}^\perp \Xdata^{(H+1)} \Phi_\star^\top \Fdata_\star^\top}_F^2.
\end{align*}
Plugging this in above, we have that with probability at least $1-\frac{2\delta}{5}$, 
\begin{align*}
    \bar \sigma_z^2 &\paren{k m H + kn \log\paren{N_1 T \frac{\bar \lambda}{\underline{\lambda}}}  + \log\paren{\frac{1}{\delta}}} \gtrsim \frac{c N_1}{N_2} \norm{P_{X^{(H+1)} \hat \Phi^\top}^\perp \Xdata^{(H+1)} \Phi_\star^\top \Fdata_\star^\top}_F^2.
\end{align*}
To conclude the proof, note that $\Xdata^{(H+1)} \hat \Phi^\top \paren{\hat F^{(H+1)}}^\top = P_{\Xdata^{(H+1)} \hat \Phi^\top} \Vdata^{(h)}$ and $\Fdata_\star^\dagger := (\Fdata_\star^\top \Fdata_\star)^{-1} \Fdata_\star^\top$ such that $\Fdata_\star^\dagger \Fdata_\star = I_k$. Thus
\begin{align*}
    & \frac{1}{N_2 T} {\norm{\Xdata^{(H+1)} \paren{\hat F^{(H+1)} \hat\Phi - F^{(H+1)}_\star \Phi_\star}^\top}_F^2} \\
    &= \frac{1}{N_2 T}{\norm{P_{\Xdata^{(H+1)} \hat \Phi^\top}^\perp \Xdata^{(H+1)} \paren{F^{(H+1)}_\star \Phi_\star}^\top + P_{\Xdata^{(H+1)} \hat \Phi^\top} \Zdata^{(H+1)}}_F^2} \\
    &\leq  \frac{1}{N_2 T}{\norm{P_{\Xdata^{(H+1)} \hat \Phi^\top}^\perp \Xdata^{(H+1)} \paren{F^{(H+1)}_\star \Phi_\star}^\top}_F^2}  + \frac{1}{N_2 T}\norm{ P_{\Xdata^{(H+1)} \hat \Phi^\top} \Zdata^{(H+1)}}_F^2 \\
    &=  \frac{1}{N_2 T}{\trace\paren{P_{\Xdata^{(H+1)} \hat \Phi^\top}^\perp \Xdata^{(H+1)}  \Phi_\star^\top \paren{F^{(H+1)}_\star} ^\top F^{(H+1)}_\star \Phi_\star \paren{\Xdata^{(H+1)}}^\top P_{\Xdata^{(H+1)} \hat \Phi^\top}^\perp }}  \\
    &\qquad \qquad + \frac{1}{N_2 T}\norm{ P_{X^{(H+1)} \hat \Phi^\top} \Zdata^{(H+1)}}_F^2 \\
    &= \frac{1}{N_2 T}{\trace\paren{P_{\Xdata^{(H+1)} \hat \Phi^\top}^\perp \Xdata^{(H+1)}  \Phi_\star^\top \Fdata_\star^\top \paren{\Fdata_\star^\dagger}^\top \paren{F^{(H+1)}_\star} ^\top F^{(H+1)}_\star \Fdata_\star^\dagger \Fdata_\star \Phi_\star \paren{\Xdata^{(H+1)}}^\top P_{\Xdata^{(H+1)} \hat \Phi^\top}^\perp }}  \\
    &\qquad \qquad + \frac{1}{N_2 T}\norm{ P_{X^{(H+1)} \hat \Phi^\top} \Zdata^{(H+1)}}_F^2 \\
    &\leq   \frac{1}{N_2T}\norm{\paren{\Fdata^\dagger_\star}^\top {\paren{F^{(H+1)}_\star} ^\top F^{(H+1)}_\star} \Fdata^\dagger_\star } \trace\paren{P_{\Xdata^{(H+1)} \hat \Phi^\top}^\perp \Xdata^{(H+1)}  \Phi_\star^\top \Fdata_\star^\top \Fdata_\star \Phi_\star \paren{\Xdata^{(H+1)}}^\top P_{\Xdata^{(H+1)} \hat \Phi^\top}^\perp }  \\
    &\qquad \qquad + \frac{1}{N_2T}\norm{ P_{\Xdata^{(H+1)} \hat \Phi^\top} \Zdata^{(H+1)}}_F^2 \\
    &\lesssim \frac{1}{N_2T} \norm{F^{(H+1)}\Fdata_\star^\dagger }^2 \norm{P_{\Xdata^{(H+1)} \hat \Phi^\top}^\perp \Xdata^{(H+1)} \Phi_\star^\top \Fdata_\star^\top}_F^2   + \frac{1}{N_2T} \norm{ P_{\Xdata^{(H+1)} \hat \Phi^\top} \Zdata^{(H+1)}}_F^2 \\
    &\lesssim \frac{1}{cN_1 T H} \bar \sigma_z^2 \paren{k n_u H + kn_x \log\paren{N_1 T \frac{\bar \lambda}{\underline{\lambda}}}  + \log\paren{\frac{1}{\delta}}}  + \frac{1}{N_2 T}\norm{ P_{\Xdata^{(H+1)} \hat \Phi^\top} \Zdata^{(H+1)}}_F^2
\end{align*}
where the last line follows from the inequality in \eqref{eq: target data projection bound} and applying Assumption~\ref{as: diverse source controllers} on $\norm{F^{(H+1)}\Fdata_\star^\dagger }^2$. As $\hat \Phi$ is independent from $\Xdata^{(H+1)}$ and $\Zdata^{(H+1)}$, the second term may be bounded by applying \Cref{prop:yasin_multiprocesses} as in Equation \eqref{eq:apply self normalized}. In particular, with probability at least $1-\frac{\delta}{5}$,
\[
    \norm{ P_{\Xdata^{(H+1)} \hat \Phi^\top} \Zdata^{(H+1)}}_F^2 \leq \sigma_z^2( k n_u + \log(1/\delta)).
\]
Therefore, 
\begin{align*}
    & \frac{1}{N_2T} \norm{\Xdata^{(H+1)} \paren{\hat F^{(H+1)} \hat\Phi - F^{(H+1)}_\star \Phi_\star}^\top}_F^2 \\
    &\lesssim \frac{1}{ cN_1 T H}  \sigma_z^2 \paren{k n_u H + kn_x \log\paren{N_1 T \frac{\bar \lambda}{\underline{\lambda}}} + \log\paren{\frac{1}{\delta}}} + \frac{\sigma_z^2 kn_u + \sigma_z^2 \log(\frac{1}{\delta})}{N_2 T} \\
    &= \sigma_z^2 \paren{ \frac{k n_u H }{ cN_1 T H} +\frac{k n_x \log\paren{N_1 T \frac{\bar \lambda}{\underline{\lambda}}} }{ cN_1 T H} +\frac{\log(1/\delta)}{ cN_1 T H} + \frac{kn_u +\log(\frac{1}{\delta})}{N_2 T}} \\
    &\lesssim  \sigma_z^2 \paren{\frac{k n_x \log\paren{N_1 T \frac{\bar \lambda}{\underline{\lambda}}} }{ cN_1 T H} + \frac{kn_u +\log(\frac{1}{\delta})}{N_2 T}}. 
\end{align*}
\end{proof}

\TargetTaskRiskBound*

\begin{proof} 
    The excess risk may be written as follows: 
    \begin{align*}
        \ER(\hat \Phi, \hat F^{(H+1)}) &= \frac{1}{2 T} \Ex_{x[0], w[0], \dots, w[T-1]} \brac{\sum_{t=0}^{T-1} \norm{(F^{(H+1)} \Phi_\star - \hat F^{(H+1)} \hat \Phi) x[t]}_2^2}  \\
        &= \frac{1}{2 T} \Ex_{x[0], w[0], \dots, w[T-1]} \brac{\sum_{t=0}^{T-1} \trace\paren{\paren{F^{(H+1)} \Phi_\star - \hat F^{(H+1)} \hat \Phi} x[t] x[t]^\top \paren{F^{(H+1)} \Phi_\star - \hat F^{(H+1)} \hat \Phi}} } \\
        &= \frac{1}{2 T} \sum_{t=0}^{T-1} \trace\paren{\paren{F^{(H+1)} \Phi_\star - \hat F^{(H+1)} \hat \Phi} \Sigma_x^{(H+1)} \paren{F^{(H+1)} \Phi_\star - \hat F^{(H+1)} \hat \Phi}}  \\
        &= \frac{1}{2 } \trace\paren{\paren{F^{(H+1)} \Phi_\star - \hat F^{(H+1)} \hat \Phi} \Sigma_x^{(H+1)} \paren{F^{(H+1)} \Phi_\star - \hat F^{(H+1)} \hat \Phi}}.  \\
    \end{align*}
    Then by applying the inequality in Lemma~\ref{lem: target covariance concentration} with $\Phi = \bmat{\hat \Phi & \Phi_\star}$, we have 
    \begin{align*}
        0.9 \Phi^\top \Sigma_x^{(H+1)} \Phi \preceq \frac{1}{N_2T} \Phi^\top \paren{\Xdata^{(H+1)}}^\top \Xdata^{(H+1)} \Phi.
    \end{align*}
    Using this result this above, we have that
    \begin{align*}
        \ER(\hat \Phi, \hat F^{(H+1)}) &\leq \frac{1}{1.8 N_2 T} \trace\paren{\paren{F^{(H+1)} \Phi_\star - \hat F^{(H+1)} \hat \Phi} \paren{\Xdata^{(H+1)}}^\top \Xdata^{(H+1)}\paren{F^{(H+1)} \Phi_\star - \hat F^{(H+1)} \hat \Phi}} \\
        & \leq \frac{1}{1.8 N_2 T} \norm{\paren{F^{(H+1)} \Phi_\star - \hat F^{(H+1)} \hat \Phi} \Xdata^{(H+1)}}_F^2.
    \end{align*}
    The claim then follows by application of Lemma~\ref{lem: target controller concentration}.
\end{proof}

\section{Bounding the imitation gap}\label{appendix: bounding imitation gap}
We recall that the issue with translating a bound on the excess risk of the target task $\ER(\hat \Phi, \hat F^{(H+1)})$ into a bound on the tracking error between the closed-loop learned and expert trajectories comes from the fundamental distribution shift between the expert trajectories seen during training and the trajectories generated by running the learned controller in closed-loop. This has traditionally been a difficult problem to analyze due to the circular nature of requiring the feedback controller error to be small to guarantee small state deviation, which depends on the state deviation being small to begin with. 

Toward addressing this issue, recent work by \cite{pfrommer2022tasil} proposes a theoretical framework for non-linear systems to bound the tracking error by the imitation error via matching the higher-order input/state derivatives $D^p_x \pi(x)$ of the expert controller. For linear systems, matching the Jacobian is sufficient, where we note the Jacobian of a linear controller $u[t] = Kx[t]$ is simply $K$. Furthermore, a generalization bound on the empirical risk of a learned controller such as \Cref{thm: target task excess risk bound} implicitly bounds the controller error $\norm{\hat K - K^{(H+1)}}$. 

However, the framework described in \cite{pfrommer2022tasil} crucially relies on assuming a noiseless system, whereas the problem considered in this work is only non-trivial in the presence of process noise. Furthermore, due to the lack of excitatory noise and the generality of non-linear systems, the tracking error bounds in \cite{pfrommer2022tasil} scale with $\tilde\calO\paren{\frac{\mathrm{\# param}}{N_2 \delta'}}$, where $\delta'$ is the failure probability over a new trajectory. To that end, the main goal of this section is to introduce bounds on the tracking error that scale multiplicatively with the favorable generalization bounds on the excess risk, which improve with the trajectory length $T$, and also improving the system-agnostic Markov scaling $1/\delta'$ to $\log(1/\delta')$ in our linear systems setting.

Toward proving a bound on the imitation gap, we first introduce the notion of general stochastic incremental stability ($\delta$-SISS). We recall definitions of standard comparison functions \citep{khalil1996nonlinear}: a function $\gamma(x)$ is class $\calK$ if it is continuous, strictly increasing, and satisfies $\gamma(0)=0$. A function $\beta(x, t)$ is class $\calK\calL$ if it is continuous, $\beta(\cdot, t)$ is class $\calK$ for each fixed $t$, and $\beta(x, \cdot)$ is decreasing for each fixed $x$.
\begin{definition}[$\delta$SISS]\label{def: dSISS}
Consider the discrete-time control system $x[t+1] = f(x[t],u[t],w[t])$ subject to input perturbations $\Delta[t]$ and additive zero-mean process noise $\curly{w[t]}_{t \geq 0}$. The system $f(x[t],u[t],w[t])$ is \emph{incremental stochastic input-to-state stable} ($\delta$-SISS) if for all initial conditions $x[0], y[0] \in \calX$, perturbation sequences $\curly{\Delta[0]}_{t \geq 0}$ and noise realizations $\curly{w[0]}_{t \geq 0}$ there exists a class $\calK\calL$ function $\beta$ and a class $\calK$ function $\gamma$ such that the trajectories $x[t+1] = f(x[t],u[t],w[t])$ and $y[t+1] = f(y[t],u[t]+\Delta[t], w[t])$ satisfy for all $t \in \N$
\begin{align} \label{eq: delta SISS}
    \norm{x[t] - y[t]} &\leq \beta\paren{\norm{x[0] - y[0]}, t} + \gamma\paren{\max_{0 \leq k \leq t - 1} \norm{\Delta[k]}}.
\end{align}


\end{definition}

\begin{lemma}[Stable linear dynamical systems are $\delta$SISS]\label{lem: LDS are EdISS}
    Let $(A,B)$ describe a linear time-invariant system $x[t+1] = Ax[t] + B u[t]$. Let $A$ be Schur-stable, i.e.\ $\rho(A) < 1$, where $\rho(\cdot)$ denotes the spectral radius. 
    Recall the constant $\calJ(A) := \sum_{t\geq 0} \norm{A^t}$. Furthermore, for any given $\nu$ such that $\rho(A) < \nu < 1$, define the corresponding constant
    \begin{align*}
        \tau(A,\nu) := \sup_{k \in \N} \frac{\norm{A^k}}{\nu^k}, 
    \end{align*}
    which is guaranteed to be finite by Gelfand's formula \citep[Corollary 5.6.13]{horn2012matrix}. 
    The linear system $(A,B)$ is incrementally stochastic input-to-state stable, where we may set
    \begin{align*}
        \beta(x, t) &:= \tau(A,\nu) \nu^t x \\
        \gamma(x) &:= \calJ(A) \norm{B} x.
    \end{align*}
\end{lemma}
\begin{proof}
Let us define the nominal and perturbed states $x_t$ and $y_t$ defined by
\begin{align*}
    x[t+1] &= Ax[t] + Bu[t] + w[t], \quad x[0] = \xi_1 \\
    y[t+1] &= Ay[t] + B(u[t] + \Delta[t]) + w[t], \quad y[0] = \xi_2.
\end{align*}
We observe that by linearity, we may write
\begin{align*}
    x[t] &= A^t x[0] + \sum_{k=0}^{t-1} (A^{t-k-1} B u[k] + A^{t-k-1} w[k]) \\
    y[t] &= A^t y[0] + \sum_{k=0}^{t-1} (A^{t-k-1} B u[k] + A^{t-k-1} B \Delta[k] + A^{t-k-1} w[k]).
\end{align*}
Therefore, their difference can be written as
\begin{align*}
    x[t] - y[t] &= A^t(x[0] - y[0]) - \sum_{k=0}^{t-1} A^{t-k-1} B \Delta[k] \\
    \implies \norm{x[t] - y[t]} &\leq \norm{A^t} \norm{x[0] - y[0]} + \paren{\sum_{k=0}^{t-1} \norm{A^{t-k-1}}}\norm{B} \max_{0 \leq k \leq t-1} \norm{\Delta[k]}  \\
    &\leq \tau(A,\nu) \nu^t \norm{x[0] - y[0]} + \calJ(A) \norm{B} \max_{0 \leq k \leq t-1} \norm{\Delta[k]}
\end{align*}
where we can match the functions $\beta(\cdot, t)$ and $\gamma(\cdot)$ to their respective quantities above. Crucially, we observe that due to linearity, the noise terms in $x[t]$ and $y[t]$ cancel out, and therefore $\delta$SISS is effectively the same as the standard $\delta$ISS.
\end{proof}

By showing stable linear systems are $\delta$SISS, we may adapt Corollary A.1 from \cite{pfrommer2022tasil} to yield the following guarantee with respect to the expert closed-loop system of the target task $A + BK^{(H+1)}$, which allows us to bound the imitation gap in terms of the maximal deviation between the learned and target controller inputs.
\begin{proposition}\label{prop: linear TaSIL guarantee}
Let the target task closed-loop system $(A+BK^{(H+1)}, B)$ be $\delta$SISS for $\beta(\cdot, t)$ and $\gamma(\cdot)$ defined in \Cref{lem: LDS are EdISS}. For a given test controller $K$, and realizations of randomness $x[0] \sim \calN(0, \Sigma_x^{(H+1)})$, $z[t] \sim \calN(0, \sigma_z^2 I)$, $w[t] \sim N(0, \Sigma_w^{(H+1)}), t = 0, \dots, T-1$, we write
\begin{align}\label{eq: identical coupling}
\begin{split}
    x_\star[t+1] &= (A + BK^{(H+1)}) x_\star[t] + Bz[t] + w[t], \quad x_\star[0] = x[0] \\
    \hat x[t+1] &= (A + B K)\hat x[t] + Bz[t] + w[t],  \quad \hat x[0] = x[0].
\end{split}
\end{align}
In other words, $x_\star[t]$ is the state from rolling out the expert target task controller $K^{(H+1)}$, and $\hat x[t]$ is the state from rolling out the test controller $K$ under the same conditions. If $K$ satisfies
\begin{align*}
        \norm{K-K^{(H+1)}} \leq \frac{1}{2\calJ(A+BK^{(H+1)})\norm{B}},
\end{align*}
then the tracking error satisfies for any $T \geq 1$,
\begin{align*}
    \max_{1 \leq t \leq T} \norm{\hat x[t] - x_\star[t]} \leq \max_{0 \leq t \leq T-1} 2\calJ(A+BK^{(H+1)})\norm{B} \norm{\paren{K  - K^{(H+1)} } x_\star[t]}.
\end{align*}

\end{proposition}

\noindent In other words, Proposition~\ref{prop: linear TaSIL guarantee} allows us to bound the imitation gap by the excess risk induced by controller $K$, as long as $K$ is sufficiently close to the expert controller $K^{(H+1)}$ in the spectral norm.

\begin{proof}
We borrow Proposition 3.1 from \cite{pfrommer2022tasil}, which states a general result for non-linear controllers and non-linear systems, but in the noiseless setting, and specify it to the linear systems setting with process noise. In particular, the result states that for a given state-feedback controller $\pi(x)$ and an expert state-feedback controller $\pi_\star(x)$ that induces a $\delta$-SISS closed-loop system $x_\star[t+1] = f(x_\star[t], \pi_\star(x_\star[t])) + w[t]$ (the result can be extended from $\delta$-ISS with no modification to the proof), then for a given $\varepsilon > 0$, if $\pi$ satisfies the following on an expert trajectory generated by realizations of randomness $x[0], w[0], \dots, w[T-1]$:
\begin{align}\label{eq: prop 3.1 tasil}
    \max_{0 \leq t \leq T-1} \sup_{\norm{\delta} \leq \varepsilon} \norm{\pi_\star(x_\star[t] + \delta) - \pi(x_\star[t] + \delta)} &\leq \gamma^{-1}(\varepsilon),
\end{align}
where $\gamma^{-1}$ is the (generalized) inverse of the class-$\calK$ function $\gamma(\cdot)$, then the imitation gap is bounded by
\[
\max_{0 \leq t \leq T-1} \norm{\hat x[t] - x_\star[t]} \leq \varepsilon,
\]
where $\hat x[t]$ are the states generated by running controller $\pi$ in closed-loop on the same noise realizations as those generating $x_\star[t]$. Now, substituting $\pi(x) := Kx$, $\pi_\star(x) := K^{(H+1)} x$, and $\gamma(x) := \calJ(A+BK^{(H+1)})\norm{B}x$, such that $\gamma^{-1}(x) = \paren{\calJ(A+BK^{(H+1)})\norm{B}}^{-1}x$, we observe
\begin{align*}
    &\max_{0 \leq t \leq T-1} \sup_{\norm{\delta} \leq \varepsilon} \norm{\pi_\star(x_\star[t] + \delta) - \pi(x_\star[t] + \delta)} \\
    = &\max_{0 \leq t \leq T-1} \sup_{\norm{\delta} \leq \varepsilon} \norm{(K - K^{(H+1)})x_\star[t] + (K - K^{(H+1)})\delta} \\
    \leq &\max_{0 \leq t \leq T-1} \norm{(K - K^{(H+1)})x_\star[t]} + \norm{K - K^{(H+1)}} \varepsilon.
\end{align*}
Therefore, in order to satisfy \eqref{eq: prop 3.1 tasil}, it suffices to satisfy
\begin{align*}
    \max_{0 \leq t \leq T-1} \norm{(K - K^{(H+1)})x_\star[t]} + \norm{K - K^{(H+1)}} \varepsilon \leq \paren{\calJ(A+BK^{(H+1)})\norm{B}}^{-1} \varepsilon.
\end{align*}
Since $\varepsilon > 0$ is arbitrary, setting $\varepsilon := \max_t 2 \calJ(A+BK^{(H+1)})\norm{B} \norm{(K - K^{(H+1)})x_\star[t]}$, it is sufficient for
\[
\norm{K - K^{(H+1)}} \leq \frac{1}{2\calJ(A+BK^{(H+1)})\norm{B}},
\]
to satisfy the above inequality, which leads to the following bound on the imitation gap
\begin{align*}
    \max_{0 \leq t \leq T-1} \norm{\hat x[t] - x_\star[t]} &\leq \varepsilon = \max_{0 \leq t \leq T-1} 2\calJ(A+BK^{(H+1)})\norm{B} \norm{K x_\star[t] - K^{(H+1)}x_\star[t]}.
\end{align*}
This completes the proof.
\end{proof}

Before moving on, we discuss a few qualitative properties of Proposition~\ref{prop: linear TaSIL guarantee}. First off, a high-probability bound on the tracking error is somewhat higher resolution than the LQR-type bounds in \cite{fazel2018global, mania2019certainty}, where the quantity of interest is the difference in the expected infinite-horizon costs of the two controllers, which does not directly imply a bound on the deviation between the states induced by the two controllers at a given time. Furthermore, we note Proposition~\ref{prop: linear TaSIL guarantee} is meaningful \textit{precisely} because there is driving process noise. If the target task is noiseless, and test controller $K$ is stabilizing, then $\norm{\hat x[t] - x_\star[t]}$ is trivially upper bounded by an exponentially decaying quantity, which for a sufficiently long horizon $T$ will beat out any generalization bound scaling with $\mathrm{poly}(N_1,N_2, H, T)$--however, this noiseless setting is correspondingly uninteresting to study as a statistical problem.

We propose the following key result in the form of a high probability bound on the tracking error.

\begin{theorem}[Full version of \Cref{thm: final imitation gap bound short}, \eqref{eq: Thm 3.2 high prob bound}]\label{thm: final imitation gap bound full}
    Let the target task closed-loop system $(A+BK^{(H+1)}, B)$ be $\delta$SISS for $\beta(\cdot, t)$ and $\gamma(\cdot)$ as defined in \Cref{lem: LDS are EdISS}.
    Given a generalization bound on the excess risk of the learned representation and target task weights $(\hat\Phi, \hat F^{(H+1)})$ (such as \Cref{thm: target task excess risk bound}) of the form: with probability greater than $1 - \delta$
    \begin{align*}
        \ER\paren{\hat\Phi, \hat F^{(H+1)}} &\leq f(N_1, T, H, N_2, \delta),
    \end{align*}
    we have the following bound on the tracking error.
    Assuming we have enough samples such that
    \begin{equation*}
        f(N_1, T, H, N_2, \delta) \leq \frac{\lambda_{\min}(\Sigma_x^{(H+1)})}{8\calJ(A+BK^{(H+1)})^2\norm{B}^2},
    \end{equation*}
    then with probability greater than $1- \delta - \delta'$, for a new target task trajectory sampled with i.i.d.\ process randomness $x[0] \sim \Sigma_x^{(H+1)}$, $w[t] \sim \Sigma_w^{(H+1)}$, $t = 0, \dots, T-1$, the tracking error satisfies
    \begin{align}
        \max_{1 \leq t \leq T} \norm{\hat x[t] - x_\star[t]}^2 &\leq 4\calJ(A+BK^{(H+1)})^2\norm{B}^2 \paren{1 +  4 \log\paren{\frac{T}{\delta'}}} \mathrm{ER}(\hat{\Phi}, \hat W^{H+1}) \\
        &\lesssim \calJ(A+BK^{(H+1)})^2\norm{B}^2 \log\paren{\frac{T}{\delta'}} f(N_1, T, H, N_2, \delta),
    \end{align}
    where the expert and learned trajectory states $\hat x[t]$ and $x_\star[t]$ are as defined in \Cref{prop: linear TaSIL guarantee}.
    
\end{theorem}

\begin{proof}
By Proposition~\ref{prop: linear TaSIL guarantee}, we have that if the learned controller $\hat K = \hat F^{(H+1)} \hat \Phi$ satisfies
\begin{align*}
    \norm{\hat K - K^{(H+1)}} \leq \frac{1}{2\calJ(A+BK^{(H+1)})\norm{B}},
\end{align*}
where we plugged in the $\beta(\cdot, t)$, $\gamma(\cdot)$ functions derived from Lemma~\ref{lem: LDS are EdISS} on the target task expert closed-loop system $(A+BK^{(H+1)}, B)$, then we have for learned and expert trajectories generated by independent instances of randomness $x[0], w[0], \dots, w[T-1]$ the following imitation gap:
\[
\max_{1 \leq t \leq T} \norm{\hat x[t] - x_\star[t]} \leq \max_{0 \leq t \leq T-1} 2\calJ(A+BK^{(H+1)})\norm{B} \norm{\paren{\hat K - K^{(H+1)}}x_\star[t]}.
\]
In order derive sample complexity bounds from this, we observe a generalization bound on the excess risk
\begin{align*}
        \ER(\hat \Phi, \hat F^{(H+1)}) &:= \frac{1}{2}\Ex_{\curly{x_\star[t]} \sim \calP^{(H+1)}} \brac{\frac{1}{T} \sum_{t=0}^{T-1} \norm{ \paren{\hat K - K^{(H+1)}} x_\star[t] }^2 } \\
        &= \frac{1}{2T}\sum_{t=0}^{T-1} \Ex\brac{\trace\paren{(\hat K  - K^{(H+1)})x[t]x[t]^\top (\hat K - K^{(H+1)})^{\top}}} \\
        &= \frac{1}{2} \trace\paren{(\hat K - K^{(H+1)})\Sigma_x^{(H+1)} (\hat K - K^{(H+1)})^{\top}} \\
        &\leq  f(N_1, T, H, N_2, \delta) \quad \text{w.p. }\geq 1-\delta
\end{align*}
directly implies a generalization bound on the Frobenius norm deviation between the learned and expert controllers:
\begin{align*}
        &\frac{1}{2} \trace\paren{(\hat K - K^{(H+1)})\Sigma_x^{(H+1)} (\hat K - K^{(H+1)})^{\top}} \geq \frac{1}{2} \lambda_{\min}(\Sigma_x^{(H+1)}) \norm{\hat K^{(H+1)} - K^{(H+1)} }_F^2 \\
        \implies &\norm{\hat K^{(H+1)} - K^{(H+1)} }_F^2 \leq \frac{2}{\lambda_{\min}(\Sigma_x^{(H+1)})} f(N_1, T, H, N_2, \delta)\text{ w.p. }\geq 1-\delta.
\end{align*}
Since the Frobenius norm upper bounds the spectral norm, it suffices to have enough samples $N_1, N_2, T, H$ such that
\[
f(N_1, T, H, N_2, \delta) \leq \frac{\lambda_{\min}(\Sigma_x^{(H+1)})}{8\calJ(A+BK^{(H+1)})^2\norm{B}^2}
\]
If the sample complexity requirement is satisfied, then we have with probability greater than $1 - \delta$ that the spectral norm gap is satisfied
\begin{align*}
    \norm{\hat K - K^{(H+1)}} \leq \frac{1}{4\calJ(A+BK^{(H+1)})^2\norm{B}^2},
\end{align*}
which in turn implies the bound on the tracking error
\begin{align}\label{eq: imitation error to tracking error}
\max_{1 \leq t \leq T} \norm{\hat x[t] - x_\star[t]}^2 \leq \max_{0 \leq t \leq T-1} 4\calJ(A+BK^{(H+1)})^2\norm{B}^2 \norm{\paren{\hat K - K^{(H+1)}}x_\star[t]}^2.
\end{align}
In order to convert the RHS of the above expression, which involves a maximum over time of the imitation error between $\hat K$ and $K^{(H+1)}$, into something involving the excess risk of the controller $\hat K$, which is the expected value of the imitation error, we again appeal to the Hanson-Wright inequality, adapted specifically for Gaussian quadratic forms in \Cref{prop: Hanson-Wright for Gaussians}.
We recall from \eqref{eq: imitation error to tracking error} the tracking error is with probability greater than $1 - \delta$ bounded by
\[
\max_{1 \leq t \leq T} \norm{\hat x[t] - x_\star[t]}^2 \leq \max_{0 \leq t \leq T-1} 4\calJ\paren{A + BK^{(H+1)}}^2 \norm{B}^2\norm{\paren{\hat K - K^{(H+1)}}x_\star[t]}^2.
\]
In order to bound the RHS of the above inequality by a multiplicative factor of the excess risk $\ER(\hat \Phi, \hat F^{(H+1)})$, we apply \Cref{prop: Hanson-Wright for Gaussians}, setting
\begin{align}\label{eq: whitening}
    R := \paren{\hat K - K^{(H+1)}} \paren{\Sigma_x^{(H+1)}}^{1/2}, \quad z_t := \paren{\Sigma_x^{(H+1)}}^{-1/2} x_\star[t].
\end{align}
By an application of the union bound, we have
\begin{align*}
    \prob\brac{\max_{t \leq T-1} z_t^\top R^\top R z_t \geq (1+\varepsilon) \norm{R}_F^2 } &\leq \sum_{t=0}^{T-1} \prob\brac{z_t^\top R^\top R z_t \geq (1+\varepsilon) \norm{R}_F^2 }\\
    &\leq T  \exp\paren{- \frac{1}{4} \min\curly{\frac{\varepsilon^2}{4}, \varepsilon } \frac{\norm{R}_F^2}{\norm{R}^2} }.
\end{align*}
Setting the last line less than the desired failure probability $\delta' \in (0,1)$, we get
\begin{align*}
    \min\curly{\frac{\varepsilon^2}{4}, \varepsilon } &\geq 4\frac{\norm{R}^2}{\norm{R}_F^2}\log\paren{\frac{T}{\delta'}}.
\end{align*}
Since $\norm{R} \leq \norm{R}_F$, it suffices to choose
\begin{align*}
    \varepsilon &= 4 \log\paren{\frac{T}{\delta'}},
\end{align*}
such that with probability greater than $1 - \delta'$
\begin{align*}
    \max_{0 \leq t \leq T-1} \norm{\paren{\hat K - K^{(H+1)}}x_\star[t]}^2 &\leq \paren{1 + 4 \log\paren{\frac{T}{\delta'}}} \norm{\paren{\hat K - K^{(H+1)}} \paren{\Sigma_x^{(H+1)}}^{1/2}}_F^2 \\
    &= \paren{1 + 4 \log\paren{\frac{T}{\delta'}}} \ER\paren{\hat\Phi, \hat F^{(H+1)}}.
\end{align*}
Plugging this back into the tracking error bound and union bounding over the generalization bound event on $\ER\paren{\hat\phi, \hat F^{(H+1)}}$ of probability $1-\delta$ and the concentration event of probability $1-\delta'$ yields the desired result.

\end{proof}

The high probability bound introduced in \Cref{thm: final imitation gap bound full} provides a very high resolution control over the deviation between closed-loop learned and expert states, where we control the maximum deviation of states over time by the \textit{expected time-averaged} excess risk of the learned controller, accruing only a $\log(T)$ factor in the process. Furthermore, our high-probability bounds are multiplicative with respect to only $\log(1/\delta)$ rather than the naive Markov's inequality factor $1/\delta$, which is immediately higher-resolution than the corresponding in-expectation bound (for example derived from \Cref{prop: max-to-avg excess risk}), on which only Markov's inequality can be applied without more detailed analysis. However, this being said, in much of existing literature in control and reinforcement learning, the performance of a policy is evaluated as an expected cost/reward over the trajectories it generates, for example LQG cost \citep{fazel2018global}, or expected reward in online imitation learning \citep{ross2011reduction}. This therefore motivates understanding whether the tools we develop directly imply bounds in-expectation on general cost functions evaluated on the learned trajectory distribution versus the expert trajectory distribution. Intuitively, if we treat the tracking error between two trajectories as a metric, if the cost function varies continuously (e.g.\ Lipschitz) with respect to this metric, our generalization bounds should imply some sort of \textit{distributional distance} between the closed-loop learned and expert trajectory distributions. This is made explicit in the following result.


\begin{theorem}[Full version of \Cref{thm: final imitation gap bound short}, \eqref{eq: Thm 3.2 expectation bound}]\label{thm: in-expectation imitation gap full}
Let us denote stacked trajectory vectors $\vec x_{1:T} = \bmat{x[1]^\top & \cdots & x[T]^\top}^\top \in \R^{n_xT}$, and denote $\vec x^\star_{1:T} \sim \calP^\star_{1:T}$ and $\hat{\vec x}_{1:T} \sim \hat\calP_{1:T}$ as the distributions of closed-loop expert and learned trajectories generated by $K^{(H+1)}$ and $\hat K$, respectively. Let $h$ be any cost function that is $L$-Lipschitz with respect to the metric between trajectories $d\paren{\vec x_{1:T}, \vec y_{1:T}} = \max_{1 \leq t \leq T} \norm{x[t] - y[t]}$, i.e.,
\begin{align*}
    \abs{h(x) - h(y)} \leq Ld(x,y), \;\forall x,y \in \calX
\end{align*}
Assume the sample requirements of \Cref{thm: final imitation gap bound full} are satisfied for given $\delta$. Then with probability greater than $1 - \delta$, the following bound holds on the gap between the expected costs of expert and learned trajectories:
\begin{align*}
    \abs{\Ex_{\hat\calP_{1:T}}\brac{h(\hat{\vec x}_{1:T})} - \Ex_{\calP^\star_{1:T}}\brac{h(\vec x^\star_{1:T})}} &\leq 2\sqrt{3} L\calJ(A_{\mathsf{cl}})\norm{B} \sqrt{1+\log(T)} \sqrt{\mathrm{ER}(\hat{\Phi}, \hat F^{(H+1)})}
\end{align*}
where $A_{\mathsf{cl}} := A+BK^{(H+1)}$.

\end{theorem}

\begin{remark}
We note that since $\frac{1}{T}\sum_{t=1}^{T} \norm{x[t] - y[t]} \leq \max_{1 \leq t \leq T} \norm{x[t] - y[t]}$, the above result holds with minimal modification for $d(\vec x_{1:T}, \vec y_{1:T}) = \frac{1}{T}\sum_{t=1}^{T} \norm{x[t] - y[t]}$.
\end{remark}

Before proceeding with the proof of \Cref{thm: in-expectation imitation gap full}, the following are valid examples of $h(\cdot)$:
\begin{itemize}
    \item $h(\vec x_{1:T}) = \max_{1 \leq t \leq T} \norm{Q^{1/2} x[t]}$,
    \item $h(\vec x_{1:T}) = \max_{1 \leq t \leq T} \norm{x[t] - x_{\mathrm{goal}}[t]} + \lambda \norm{Rx[t]}$.
\end{itemize}

\begin{proof}
We appeal to an application of Kantorovich-Rubinstein duality to establish general in-expectation guarantees. Let us define the metric
\begin{align*}
    d\paren{\vec x_{1:T}, \vec y_{1:T}} &= \max_{1 \leq t \leq T} \norm{x[t] - y[t]}.
\end{align*}
Define $\Gamma(\calP^\star_{1:T}, \hat\calP_{1:T})$ as the set of all couplings between the distributions $\calP^\star_{1:T}$ and $\hat\calP_{1:T}$. In particular, trajectories $\vec x^\star_{1:T}, \hat{\vec x}_{1:T}$ following the same instances of $w[t]$ and $z[t]$ considered in \eqref{eq: identical coupling} is one such coupling. The Wasserstein $1$-distance between $\calP^\star_{1:T}$ and $\hat\calP_{1:T}$ is defined as
\begin{align*}
    \calW_1\paren{\calP^\star_{1:T}, \hat\calP_{1:T}} &= \inf_{\gamma \in \Gamma(\calP^\star_{1:T}, \hat\calP_{1:T}) }   \Ex_{\paren{\vec x^\star_{1:T}, \hat{\vec x}_{1:T}} \sim \gamma}\brac{d\paren{\vec x^\star_{1:T}, \hat{\vec x}_{1:T}}} \\
    &= \inf_{\gamma \in \Gamma(\calP^\star_{1:T}, \hat\calP_{1:T}) }   \Ex_{\paren{\vec x^\star_{1:T}, \hat{\vec x}_{1:T}} \sim \gamma}\brac{\max_{1 \leq t \leq T} \norm{x_\star[t] - \hat x[t]}}.
\end{align*}
In particular, defining the coupling \eqref{eq: identical coupling} as $\overline{\gamma}$, we immediately have
\begin{align*}
    \calW_1\paren{\calP^\star_{1:T}, \hat\calP_{1:T}} &\leq  \Ex_{\paren{\vec x^\star_{1:T}, \hat{\vec x}_{1:T}} \sim \overline{\gamma}}\brac{\max_{1 \leq t \leq T} \norm{x_\star[t] - \hat x[t]}}.
\end{align*}
Kantorovich-Rubinstein duality (cf.\ \cite{villani2021topics}) provides the following dual characterization of the Wasserstein distance:
\begin{align*}
    \frac{1}{L}\sup_{\norm{h}_{\mathrm{Lip}} \leq L}  \Ex_{\hat\calP_{1:T}}\brac{h(\hat{\vec x}_{1:T})} - \Ex_{\calP^\star_{1:T}}\brac{h(\vec x^\star_{1:T})}  &= \calW_1\paren{\calP^\star_{1:T}, \hat\calP_{1:T}},
\end{align*}
where
\[
\norm{h}_{\mathrm{Lip}} = \sup_{x,y \in \calX} \frac{\abs{h(x) - h(y)}}{d(x,y)}.
\]
Therefore, given any cost function $h$ that is $L$-Lipschitz continuous with respect to $d(x,y)$, the gap between the expected closed-loop costs of the learned and expert controllers, we can chain inequalities to yield
\begin{align*}
    \Ex_{\hat\calP_{1:T}}\brac{h(\hat{\vec x}_{1:T})} - \Ex_{\calP^\star_{1:T}}\brac{h(\vec x^\star_{1:T})} \leq L \Ex_{\paren{\vec x^\star_{1:T}, \hat{\vec x}_{1:T}} \sim \overline{\gamma}}\brac{\max_{1 \leq t \leq T} \norm{x_\star[t] - \hat x[t]}}.
\end{align*}
It remains to provide a bound on $\Ex_{\paren{\vec x^\star_{1:T}, \hat{\vec x}_{1:T}} \sim \overline{\gamma}}\brac{\max_{1 \leq t \leq T} \norm{x_\star[t] - \hat x[t]}}$. Recalling the bound \eqref{eq: imitation error to tracking error}, this reduces to bounding the imitation error along expert trajectories $x_\star[t]$. In order to convert a bound on $\max_{1 \leq t \leq T} \norm{x_\star[t] - \hat x[t]}$ to $\max_{1 \leq t \leq T} \norm{x_\star[t] - \hat x[t]}$, we apply Jensen's inequality to yield
\begin{align*}
    \Ex\brac{\max_{1 \leq t \leq T} \norm{x_\star[t] - \hat x[t]}}^2 \leq \Ex\brac{\max_{1 \leq t \leq T} \norm{x_\star[t] - \hat x[t]}^2}
\end{align*}
The rest follows from an application of the maximal inequality.
\begin{proposition}\label{prop: max-to-avg excess risk}
Given a test controller $K$, we have the following maximal-type inequality between the excess maximal risk and the excess average risk over the expert target task data:
\begin{align*}
    \Ex\brac{\max_{0\leq t \leq T-1} \norm{(K - K^{(H+1)})x_{\star}[t]}^2  } &\leq 3(1+\log(T))\Ex\brac{\frac{1}{T} \sum_{t=0}^{T-1} \norm{(K - K^{(H+1)})x_{\star}[t]}^2 }.
\end{align*}

\end{proposition}

\begin{proof}
We recall that the population distribution of $x_{\star}[t]$ is $\calN(0,\Sigma_x^{(H+1)})$. Recalling the definitions in \eqref{eq: whitening}, we can re-write 
\begin{align*}
    \norm{(K - K^{(H+1)})x_{\star}[t]}^2 &= z_t^\top R^\top R z_t,
\end{align*}
where $z_t \sim \calN(0, I)$.
Therefore, what we need to show is a maximal inequality on
\begin{align*}
    \Ex\brac{\max_{0 \leq t \leq T-1} z_t^\top R^\top R z_t}.
\end{align*}
Toward establishing such an inequality, we prove the following bound on the moment-generating function of $z_t^\top R^\top R z_t$.
\begin{lemma}\label{lem: quad form MGF bound}
    Let $z \sim \calN(0,I_n)$ be a standard Gaussian random vector, and $R \in \R^{d \times n}$ is an arbitrary fixed matrix. Then, we have the following bound on the moment-generating function of $z^\top R^\top R z$
    \begin{align*}
        \Ex\brac{\exp\paren{\lambda z^\top R^\top R z}} &\leq \exp\paren{3 \lambda \norm{R}_F^2},\quad \text{for }\abs{\lambda} \leq \frac{1}{
        3\norm{R}^2}.
    \end{align*}
    
\end{lemma}

\begin{proof}
The proof largely follows standard results, for example from \citet[Chapter 6.2]{vershynin2018high}, but we derive the result from scratch to preserve explicit numerical constants. 
By the rotational invariance of Gaussian random vectors, we may write
\begin{align*}
    \Ex\brac{\exp\paren{\lambda z^\top R^\top R z}} &= \Ex\brac{\exp\paren{\sum_{i=1}^{\min\curly{n,d}} \lambda \sigma_i^2 g_i^2}} \\
    &= \prod_{i=1}^{\min\curly{n,d}} \Ex\brac{\exp\paren{\lambda \sigma_i^2 g_i^2}},
\end{align*}
where $\sigma_i$ is the $i$th singular value of $R$ and $g_i \sim \calN(0,1)$ are i.i.d.\ standard Gaussian random variables. 
Thus we have reduced the problem to bounding the MGF of a $\chi^2(1)$ random variable. We recall that the MGF of a $\chi^2(1)$ random variable is given by
\[
\Ex\brac{\exp\paren{t g_i^2 }} = \frac{1}{\sqrt{1 - 2t}}, \quad t<\frac{1}{2}.
\]
We claim that there exists constants $C,c>0$ such that
\[
\Ex\brac{\exp\paren{t g_i^2 }} \leq \exp\paren{Ct}, \quad t \leq c <\frac{1}{2}.
\]
We can derive candidates for $C,c$ by solving
\begin{align*}
    \frac{1}{\sqrt{1 - 2t}} &\leq \exp\paren{Ct} \;\land \; t < 1/2 \\
    \iff -\frac{1}{2}\log(1-2t) &\leq Ct \; \land \; t< 1/2.
\end{align*}
We observe the function $f(t) :=  -\frac{1}{2}\log(1-2t) - Ct$ satisfies $f(0) = 0$ and $f'(t) = \frac{1}{1 - 2t} - C$ is monotonically increasing for $t < 1/2$, and $f'(0) < 0$ for $C > 1$. Therefore, for a given $C > 1$ it suffices to find $0<c<1/2$ such that $f'(c) = 0$, which implies $f(c) \leq 0$. Plugging in $C = 3$, $c = 1/3$ satisfies this. Therefore, we get the MGF bound
\[
\Ex\brac{\exp\paren{t g_i^2 }} \leq \exp\paren{3t}, \quad t \leq \frac{1}{3}.
\]
Applying this back on the MGF of $z^\top R^\top R z$, we have
\begin{align*}
    \Ex\brac{\exp\paren{\lambda z^\top R^\top R z}} &= \prod_{i=1}^{\min\curly{n,d}} \Ex\brac{\exp\paren{\lambda \sigma_i^2 g_i^2}} \\
    &\leq \prod_{i=1}^{\min\curly{n,d}} \exp\paren{3 \lambda \sigma_i^2}, \quad \lambda \leq \frac{1}{3\sigma_i^2} \; \forall i \\
    &= \exp\paren{3\lambda \sum_{i}\sigma_i^2}, \quad\lambda \leq \frac{1}{3\sigma_i^2} \; \forall i \\
    &= \exp\paren{3\lambda \norm{R}_F^2},\quad \lambda \leq \frac{1}{3\norm{R}^2},
\end{align*}
which completes the proof.
\end{proof}

Now, leveraging \Cref{lem: quad form MGF bound}: for $\lambda > 0$, we write
\begin{align*}
    \exp\paren{\lambda \Ex\brac{\max_{t=0,\dots,T-1} z_t^\top R^\top R z_t} } &\leq \Ex\brac{\exp\paren{\lambda \max_t z_t^\top R^\top R z_t}} && \text{Jensen's}\\
    &= \Ex\brac{\max_t \exp\paren{\lambda z_t^\top R^\top R z_t}} && \text{monotonicity} \\
    &\leq \sum_{t=0}^{T-1} \Ex\brac{\exp\paren{\lambda z_t^\top R^\top R z_t }} && 
\end{align*}
We note that in the last line, we are taking the expectation of each $z_t$ over its population distribution, such that $z_t \sim \calN(0,I)$. Thus,
\begin{align*}
    \exp\paren{\lambda \Ex\brac{\max_{t=0,\dots,T-1} z_t^\top R^\top R z_t} } &\leq \sum_{t=0}^{T-1} \exp\paren{3\lambda \norm{R}_F^2} = T \exp\paren{3\lambda \norm{R}_F^2}, \quad \lambda \leq \frac{1}{3\norm{R}^2}.
\end{align*}
Taking the logarithm of both sides, re-arranging, and taking the infimum over $\lambda$, we have
\begin{align*}
    \Ex\brac{\max_{t=0,\dots,T-1} z_t^\top R^\top R z_t}  &\leq \inf_{\lambda \in \Big[0, \frac{1}{3\norm{R}^2}\Big]} \frac{\log(T)}{\lambda} + 3 \norm{R}_F^2 \\
    &= 3 \norm{R}^2 \log(T) + 3\norm{R}_F^2 \\
    &\leq 3\paren{1 + \log(T)} \norm{R}_F^2.
\end{align*}
Substituting back $R = (K - K^{(H+1)})\paren{\Sigma_x^{(H+1)}}^{1/2}$, and observing
\begin{align*}
    \norm{R}_F^2 = \norm{(K - K^{(H+1)})\paren{\Sigma_x^{(H+1)}}^{1/2}}_F^2 &= \trace\paren{(K - K^{(H+1)})\Sigma_x^{(H+1)}(K - K^{(H+1)})^\top} \\
    &= \frac{1}{T}\sum_{t=0}^{T-1} \Ex\brac{\norm{Kx_{\star}[t] - K^{(H+1)}x_{\star}[t]}^2},
\end{align*}
the proof of Proposition~\ref{prop: max-to-avg excess risk} is complete.
\end{proof}

\noindent With the maximal inequality in hand, and defining $A_{\mathsf{cl}} := A+BK^{(H+1)}$, we may now bound:
\begin{align*}
    &\Ex_{\hat\calP_{1:T}}\brac{h(\hat{\vec x}_{1:T})} - \Ex_{\calP^\star_{1:T}}\brac{h(\vec x^\star_{1:T})} \\ \leq\;& L \Ex_{\paren{\vec x^\star_{1:T}, \hat{\vec x}_{1:T}} \sim \overline{\gamma}}\brac{\max_{1 \leq t \leq T} \norm{x_\star[t] - \hat x[t]}} \\
    \leq\;& L \sqrt{\Ex_{\paren{\vec x^\star_{1:T}, \hat{\vec x}_{1:T}} \sim \overline{\gamma}}\brac{\max_{1 \leq t \leq T} \norm{x_\star[t] - \hat x[t]}^2} } \\
    \leq\;& L \sqrt{\Ex_{\calP^\star_{0:T-1}}\brac{\max_{0 \leq t \leq T-1} 4\calJ(A_{\mathsf{cl}})^2\norm{B}^2 \norm{\paren{\hat K - K^{(H+1)}}x_\star[t]}^2}} \\
    \leq\;& 2\sqrt{3} L\calJ(A_{\mathsf{cl}})\norm{B} \sqrt{1+\log(T)} \sqrt{\Ex_{\calP^\star_{0:T-1}}\brac{\frac{1}{T} \sum_{t=0}^{T-1} \norm{(K - K^{(H+1)})x_{\star}[t]}^2 }} \\
    \leq\;& 2\sqrt{3} L\calJ(A_{\mathsf{cl}})\norm{B} \sqrt{1+\log(T)} \sqrt{\mathrm{ER}(\hat{\Phi}, \hat F^{(H+1)})}.
\end{align*}

\end{proof}

\subsection{Tightness of Dependence on Spectral Radius in \Cref{thm: final imitation gap bound short}}
\label{appendix: lower bound spectral dependence}


Consider the following scalar LTI system
\begin{align*}
    x[t+1] = ax[t] + u[t] + w[t], \quad w[t] \sim \calN(0,1),
\end{align*}
where $a > 0$ without loss of generality. We are given the expert controller $u^\star[t] = k^\star x[t]$ that stabilizes the above system, such that we have 
\[
0 < a + k^\star < 1.
\]
Let's say our learned controller $\hat k$ attains an $\varepsilon > 0$ error with respect to $k^\star$: $\hat k = k^\star + \varepsilon$, and $a + \hat k < 1$. We recall a couple of facts: the stationary distribution induced by $k^\star$ can be derived by solving the scalar Lyapunov equation
\begin{align*}
    \sigma^2 = (a+k^\star)^2 \sigma^2 + 1 \iff \sigma^2 = \frac{1}{1 - (a+k^\star)^2}.
\end{align*}
Therefore, akin to the setting considered in our paper, we assume the initial state distribution is the expert stationary distribution. We now study the tracking between the expert and learned states evolving in the coupled manner analogous to \eqref{eq: identical coupling}
\begin{align*}
    x_\star[t+1] &= (a + k^\star)x_\star[t] + w[t], \quad x_0 \sim \calN\paren{0, \frac{1}{1 - (a+k^\star)^2} } \\
    \hat x[t+1] &= (a + k^\star + \varepsilon) \hat x[t] + w[t], \quad x_0 \sim \calN\paren{0, \frac{1}{1 - (a+k^\star)^2} }.
\end{align*}
In particular, we show the following lower bound on the tracking error
\begin{proposition}\label{prop: lower bound spectral dependence}
Given the proposed scalar system, the tracking error satisfies the following lower bound for sufficiently large $T$,
\begin{align*}
    &\Ex\brac{\max_{1 \leq t \leq T} \abs{x_\star[t] - \hat x[t]}^2} \\
    \gtrsim\; & \frac{1}{\paren{1 - (a+k^\star)}^2} \underbrace{\frac{1}{T}\Ex\brac{\sum_{t=0}^{T-1} \abs{(\hat k - k^\star)x_\star[t]}^2}}_{\text{excess risk of }\hat k \text{ on expert data}} =: \frac{1}{\paren{1 - (a+k^\star)}^2} \mathrm{ER}(\hat k).
\end{align*}
\end{proposition}

\noindent We note that instantiating the in-expectation upper bound using the maximal inequality in \Cref{prop: max-to-avg excess risk} yields
\begin{align*}
    \Ex\brac{\max_{1 \leq t \leq T} \abs{x_\star[t] - \hat x[t]}^2} &\leq 12\calJ(a+k^\star)^2 \norm{B}^2 (1+\log(T))\mathrm{ER}(\hat k) \\
    &= \frac{12}{(1 - (a+k^\star))^2} (1+\log(T))\mathrm{ER}(\hat k).
\end{align*}
Therefore, \Cref{prop: lower bound spectral dependence} states that up to a log-factor in horizon length $T$, the polynomial dependence on the expert system's spectral radius $a + k^\star$ matches that in the upper bound.

\begin{proof} We can immediately compute the excess risk in closed form
\begin{align*}
   \mathrm{ER}(\hat k) := \frac{1}{T}\Ex\brac{\sum_{t=0}^{T-1} \abs{(\hat k - k^\star)x_\star[t]}^2} &= (\hat k - k^\star)^2 \Ex\brac{\frac{1}{T}\sum_{t=0}^{T-1} {x_\star[t]}^2 } \\
    &= \frac{\varepsilon^2}{1 - (a+k^\star)^2} \quad \text{by stationarity}.
\end{align*}
Therefore, in addition to the spectral radius showing up in the excess risk, we need to show the tracking error accrues another factor of the spectral radius on top of that. We observe
\begin{align*}
    \max_{1 \leq t \leq T} \abs{x_\star[t] - \hat x[t]}^2 &\geq \abs{x_\star[{T}] - \hat x[{T}]}^2 \\
    \implies \Ex\brac{\max_{1 \leq t \leq T} \abs{x_\star[t] - \hat x[t]}^2} &\geq \max_{t \leq T} \Ex\brac{\abs{x_\star[T] - \hat x[T]}^2} \\
    &= \max_{t \leq T} \Ex\brac{{x_\star[{T}]}^2} + \Ex\brac{\hat x[{T}]^2} - 2\Ex\brac{x_\star[{T}] \hat x[{T}]} \\
    &= \max_{t \leq T} \frac{1}{1 - (a+k^\star)^2} + \frac{(a+\hat k)^{2T}}{1 - (a+k^\star)^2} + \sum_{t=0}^{T-1} (a+\hat k)^{2(T-t)} \\
    &\qquad - \frac{2(a+k^\star)^{T}(a+\hat k)^{T}}{1 - (a+k^\star)^2} -  2\sum_{t=0}^{T-1} (a+k^\star)^{T-t}(a+\hat k)^{T-t},
\end{align*}
where the summations come from the fact that the initial condition is drawn from the \textit{stationary distribution induced by the expert}, not the learned controller, in conjunction with the formula $x[t] = a^t x_0 + \sum_{k=0}^{t-1} a^{t-1-k}w[t]$ for system $x[t+1] = ax[t] + w[t]$. Since in the limit we have
\begin{align*}
    \lim_{T \to \infty} \Ex\brac{\abs{x_\star[{T}] - \hat x[{T}]}^2 } &= \frac{1}{1 - (a+k^\star)^2} + \frac{1}{1 - (a+\hat k)^2} - \frac{2}{1 - (a+k^\star)(a+\hat k)},
\end{align*}
we simply take $T$ large enough such that
\begin{align*}
    \Ex\brac{\abs{x_\star[{T}] - \hat x[{T}]}^2 } &\geq \frac{1}{2}\paren{\frac{1}{1 - (a+k^\star)^2} + \frac{1}{1 - (a+\hat k)^2} - \frac{2}{1 - (a+k^\star)(a+\hat k)}}.
\end{align*}
We now bound
\begin{align*}
&\frac{1}{1 - (a+k^\star)^2} + \frac{1}{1 - (a+\hat k)^2} - \frac{2}{1 - (a+k^\star)(a+\hat k)}\\
    =\;& \sum_{t=0}^{\infty} (a+k^\star)^{2t} + (a+\hat k)^{2t} - 2(a+k^\star)^t(a+\hat k)^t \\
    =\;& \sum_{t=0}^{\infty} ((a+\hat k)^t - (a+k^\star)^t )^2 \\
    =\;& \sum_{t=0}^{\infty} \paren{(\hat k - k^\star) \paren{(a+k^\star)^{t-1} + (a+k^\star)^{t-2}(a+\hat k) + \cdots + (a+k^\star)(a + \hat k)^{t-2} + (a + \hat k)^{t-1} }}^2 \\
    \geq\;& \varepsilon^2 \sum_{k=0}^{\infty} \paren{t (a+k^\star)^{t-1}}^2 \\
    =\;& \varepsilon^2\frac{1+(a+k^\star)^2}{(1-(a+k^\star)^2)^3},
\end{align*}
where we used the algebraic identity
\[
u^t - v^t = (u-v)(u^{t-1} + u^{t-2}v + \cdots + uv^{t-2} + v^{t-1}),
\]
and the inequality (assuming $u \leq v$)
\[
u^{t-1} + u^{t-2}v + \cdots + uv^{t-2} + v^{t-1} \geq nu^{t-1},
\]
setting $u:= a+k^\star$ and $v := a+\hat k = a+k^\star + \varepsilon$. Now recalling that the empirical risk by stationarity is given by
\[
\ER(\hat k) = \frac{\varepsilon^2}{1 - (a+k^\star)^2},
\]
we can immediately infer
\begin{align*}
    \Ex\brac{\abs{x_\star[{T}] - \hat x[{T}]}^2 } &\geq 0.5\varepsilon^2\frac{1+(a+k^\star)^2}{(1-(a+k^\star)^2)^3} \\
    &= 0.5 \ER(\hat k) \frac{1+(a+k^\star)^2}{(1-(a+k^\star)^2)^2} \\
    &> \frac{0.5}{(1-(a+k^\star)^2)^2}  \ER(\hat k).
\end{align*}
In short, we have established a lower bound on the expected imitation gap:
\begin{align*}
    \Ex\brac{\max_{0 \leq t \leq T-1} \abs{x_\star[t] - \hat x[t]}^2} &\geq \max_{t \leq T-1} \Ex\brac{\abs{x_\star[{T-1}] - \hat x[{T-1}]}^2} 
    \\
    &\gtrsim \frac{1}{(1-(a+k^\star)^2)^2} \mathrm{ER}(\hat k).
\end{align*}
We now claim that $\frac{1}{(1-(a+k^\star)^2)^2}$ matches the dependence on the spectral radius in the upper bound. Combining \Cref{prop: linear TaSIL guarantee} and \Cref{prop: max-to-avg excess risk}, we have the following upper bound on the expected tracking error:
\begin{align*}
    \Ex\brac{\max_{0 \leq t \leq T-1} \abs{x_\star[t] - \hat x[t]}^2} &\leq \frac{12}{\paren{1 - (a+k^\star)}^2} (1 + \log(T))  \mathrm{ER}(\hat k).
\end{align*}
Comparing $\frac{1}{(1-(a+k^\star)^2)^2}$ to $\frac{1}{\paren{1 - (a+k^\star)}^2}$, we get
\begin{align*}
    \frac{\frac{1}{\paren{1 - (a+k^\star)}^2}}{\frac{1}{(1-(a+k^\star)^2)^2}} &= \frac{\paren{1 - (a+k^\star)}^2 \paren{1 + (a+k^\star)}^2}{\paren{1 - (a+k^\star)}^2} = \paren{1 + (a+k^\star)}^2 \geq 1,
\end{align*}
which essentially states the dependence on the spectral radius in the lower and upper bounds match up to a constant factor:
\begin{align*}
    \frac{0.5}{(1-(a+k^\star))^2} \mathrm{ER}(\hat k) \leq \Ex\brac{\max_{0 \leq t \leq T-1} \abs{x_\star[t] - \hat x[t]}^2} \leq \frac{12}{(1-(a+k^\star))^2}(1+\log(T)) \mathrm{ER}(\hat k),
\end{align*}
which completes the result.
\end{proof}



\section{In-Expectation Bounds for LQR via the Tracking Error}\label{appendix: LQR bounds}

As previewed in \Cref{remark: Lipschitz costs}, the generalization bounds on the excess risk and tracking error are strong enough to directly imply in-expectation bounds on the LQR costs of the closed-loop expert and learned systems. However, this does not trivially follow an application of Lipschitzness with respect to the trajectory-wise metric $d(\vec x_{1:T}, \vec y_{1:T}) = \max_t \norm{x[t] - y[t]}$, due to the input cost $u^\top R u$ depending on different controllers $\hat K$ and $K^{(H+1)}$, which cannot be captured purely as a metric on trajectory states. Therefore, some massaging is required to get the analogous bounds.

Let us define the (root) LQR cost defined on $\paren{\vec x_{1:T}, K}$
\begin{align}
\begin{split}\label{eq: root LQR cost}
    h\paren{\paren{\vec x_{1:T}, K}} &:= \max_{1 \leq t \leq T} \sqrt{x[t]^\top
    \paren{Q + K^\top R K}x[t]} \\
    &= \max_{1 \leq t \leq T} \norm{\bmat{Q^{1/2} \\ R^{1/2} K} x[t]}
\end{split}
\end{align}
Our goal is to show that $h(\cdot)$ is somewhat Lipschitz with respect to the trajectory (pseudo)metric
\begin{align}
    d\paren{\paren{\vec x_{1:T}, K_1}, \paren{\vec y_{1:T}, K_2}} &:= \max_{1\leq t \leq T} \norm{x[t] - y[t]},
\end{align}
such that using the tools we developed for bounding the tracking error imply an in-expectation bound for LQR. This culminates in the following result.
\begin{proposition}\label{prop: LQR in-expectation bound}
Let 
$(\hat\Phi, \hat F^{(H+1)})$ denote the learned representation and target task weights,
and $\mathrm{ER}(\hat \Phi, \hat F^{(H+1)})$ denote the corresponding excess risk. Define $A_{\mathsf{cl}} := A+BK^{(H+1)}$. As in \Cref{thm: final imitation gap bound short}, assume that the excess risk satisfies:
\begin{equation*}
     \mathrm{ER}(\hat \Phi, \hat F^{(H+1)}) \lesssim \frac{\lambda_{\min}\paren{\Sigma_x^{(H+1)}}}{\calJ\paren{A_{\mathsf{cl}}}^2\norm{B}^2}.
\end{equation*}
Let the cost $h(\cdot)$ be the LQR cost \eqref{eq: root LQR cost}. The following in-expectation bound on the gap between the closed-loop LQR costs induced by the expert $K^{(H+1)}$ and learned controller $\hat K$ holds:
\begin{align}\nonumber
    \begin{split}
    &\abs{\Ex_{\hat\calP_{1:T}}\brac{h\paren{\paren{\vec \hat x_{1:T}, \hat K}}} - \Ex_{\calP^\star_{1:T}}\brac{h\paren{\paren{\vec x^\star_{1:T}, K^{(H+1)}}}} } \\
    \lesssim\;& \calC^{(H+1)} \sigma_z \sqrt{\log(T)} \sqrt{\frac{k n_x \log\paren{N_1 T \frac{\bar \lambda}{\underline{\lambda}}} }{ cN_1 T H} + \frac{kn_u +\log(\frac{1}{\delta'})}{N_2 T}},
    \end{split}
\end{align}
where
\[
\calC^{(H+1)} := \lambda_{\max} (Q)^{1/2} \calJ(A_{\mathsf{cl}})\norm{B} + \lambda_{\max}(R)^{1/2} \paren{\norm{K^{(H+1)}} + \sqrt{\frac{\trace\paren{\Sigma_x^{(H+1)}}}{\lambda_{\min}\paren{\Sigma_x^{(H+1)}}} } },
\]
is an expert system and LQ cost-dependent constant. 

\end{proposition}

\begin{proof} First, consider the following elementary inequality:
\begin{align*}
    \norm{M_1 x_1} - \norm{M_2 x_2} &= \norm{M_1 x_1} - \norm{M_1 x_2} + \norm{M_1 x_2} - \norm{M_2 x_2} \\
    &\leq \norm{M_1 (x_1 - x_2)} + \norm{(M_1 - M_2)x_2} \\
    &\leq \norm{M_1} \norm{x_1 - x_2} + \norm{x_2} \norm{M_1 - M_2}
\end{align*}
Applying this to $h\paren{\paren{\vec x_{1:T}, K_1}} - h\paren{\paren{\vec y_{1:T}, K_2}}$, for fixed $K_1$, $K_2$ we get
\begin{align}
    \begin{split}\label{eq: cost gap term split}
    &h\paren{\paren{\vec x_{1:T}, K_1}} - h\paren{\paren{\vec y_{1:T}, K_2}} \\
    =\;& \max_{1 \leq t \leq T} \norm{\bmat{Q^{1/2} \\ R^{1/2} K_1} x[t]} - \max_{1 \leq t \leq T} \norm{\bmat{Q^{1/2} \\ R^{1/2} K_2} y[t]} \\
    \leq\;& \max_{1 \leq t \leq T} \norm{\bmat{Q^{1/2} \\ R^{1/2} K_1} x[t]} - \norm{\bmat{Q^{1/2} \\ R^{1/2} K_2} y[t]} \\
    \leq\;& \max_{1 \leq t \leq T}  \norm{\bmat{Q^{1/2} \\ R^{1/2} K_1}} \norm{x[t] - y[t]} + \norm{y[t]} \norm{\bmat{Q^{1/2} \\ R^{1/2} K_1} - \bmat{Q^{1/2} \\ R^{1/2} K_2}} \\
    \leq\;& \norm{\bmat{Q^{1/2} \\ R^{1/2} K_1}} \max_{1 \leq t \leq T}  \norm{x[t] - y[t]} +  \lambda_{\max}(R)^{1/2} \norm{K_1 - K_2}\max_{1 \leq t \leq T}   \norm{y[t]}
    \end{split}
\end{align}
we can take the expectation of the inequality \eqref{eq: cost gap term split} to yield for any coupling of the learned and expert trajectory distributions $\Gamma(\hat\calP_{1:T}, \calP^\star_{1:T})$
\begin{align}\label{eq: cost gap bound 1}
    \begin{split}
    &\abs{\Ex_{\hat\calP_{1:T}}\brac{h\paren{\paren{\vec \hat x_{1:T}, \hat K}}} - \Ex_{\calP^\star_{1:T}}\brac{h\paren{\paren{\vec x^\star_{1:T}, K^{(H+1)}}}} } \\
    \leq\;&\lambda_{\max}\paren{Q + \hat K^\top R \hat K}^{1/2}\Ex_{\Gamma(\hat\calP_{1:T}, \calP^\star_{1:T})}\brac{\max_{t\leq T} \norm{\hat x[t] - x_\star[t] } } \\
    \quad \quad &+ \lambda_{\max}(R)^{1/2} \norm{\hat K - K^{(H+1)}}\Ex_{\calP^\star_{1:T}}\brac{\max_{1 \leq t \leq T}   \norm{x_\star[t]}}.
    \end{split}
\end{align}
Setting $\Gamma(\hat\calP_{1:T}, \calP^\star_{1:T})$ to be the coupling described in \eqref{eq: identical coupling} where the learned and expert trajectories are evaluated on the same realizations of randomness, we may apply the tools developed in \Cref{thm: in-expectation imitation gap full} to bound the first term. As for the second term, we may apply a maximal-type inequality akin to \Cref{prop: max-to-avg excess risk}. In particular, from earlier computations we have
\begin{align}\label{eq: expected tracking error}
    \Ex_{\Gamma(\hat\calP_{1:T}, \calP^\star_{1:T})}\brac{\max_{t\leq T} \norm{\hat x[t] - x_\star[t] } } &\leq 2\sqrt{3}\calJ(A_{\mathsf{cl}})\norm{B} \sqrt{1+\log(T)} \sqrt{\mathrm{ER}(\hat{\Phi}, \hat F^{(H+1)})},
\end{align}
and adapting \Cref{prop: max-to-avg excess risk} we get
\begin{align*}
\begin{split}
\Ex_{\calP^\star_{1:T}}\brac{\max_{1 \leq t \leq T}   \norm{x_\star[t]}} &\leq \sqrt{\Ex_{\calP^\star_{1:T}}\brac{\max_{1 \leq t \leq T}   \norm{x_\star[t]}^2}} \\
&\leq \sqrt{3}\sqrt{1 + \log(T)}\sqrt{\Ex\brac{\norm{x_\star[0]}^2}} \\
&\leq \sqrt{3}\sqrt{1 + \log(T)}\sqrt{\trace\paren{\Sigma_x^{(H+1)}}}.
\end{split}
\end{align*}
Lastly, from the definition of the excess risk we have
\begin{align} \label{eq: ctrl matrix difference}
\ER(\hat \Phi, \hat F^{(H+1)}) &= \frac{1}{2} \trace\paren{(\hat K - K^{(H+1)})\Sigma_x^{(H+1)} (\hat K - K^{(H+1)})^{\top}} \nonumber \\
     \implies \norm{\hat K - K^{(H+1)}} &\leq \lambda_{\min}\paren{\Sigma_x^{(H+1)}}^{-1/2} \sqrt{\ER(\hat \Phi, \hat F^{(H+1)})}.
\end{align}
Therefore, plugging expressions \eqref{eq: expected tracking error} and \eqref{eq: ctrl matrix difference} back into the bound \eqref{eq: cost gap bound 1}, we have essentially written the bound on the expected cost gap to scale with
\[
\sqrt{\log(T) \ER(\hat \Phi, \hat F^{(H+1)})},
\]
modulo the system/LQ cost-related parameters. The rest of the proof follows by instantiating the generalization bound on $\ER(\hat \Phi, \hat F^{(H+1)})$ from \Cref{thm: target task excess risk bound}, and combining the problem-related parameters. In particular, the first term of \eqref{eq: cost gap bound 1} involves the learned controller: $\lambda_{\max}\paren{Q + \hat K^\top R \hat K}$. This can be crudely upper bounded by
\begin{align*}
    \lambda_{\max}\paren{Q + \hat K^\top R \hat K}^{1/2} &\leq \lambda_{\max} (Q)^{1/2} + \lambda_{\max}(R)^{1/2} \norm{\hat K} \\
    &\leq \lambda_{\max} (Q)^{1/2} + \lambda_{\max}(R)^{1/2} \paren{\norm{K^{(H+1)}} + \frac{1}{2\calJ(A_{\mathsf{cl}})\norm{B}} },
\end{align*}
where the last line comes from reverse triangle inequality on the burn-in requirement~\eqref{eq:ER_requirement}. A tighter bound can be derived by using~\eqref{eq: ctrl matrix difference}, but the scaling of the final bound is the same. Therefore, we have
\begin{align}\nonumber
    \begin{split}
    &\abs{\Ex_{\hat\calP_{1:T}}\brac{h\paren{\paren{\vec \hat x_{1:T}, \hat K}}} - \Ex_{\calP^\star_{1:T}}\brac{h\paren{\paren{\vec x^\star_{1:T}, K^{(H+1)}}}} } \\
    \lesssim\;& \calC^{(H+1)} \sqrt{\log(T) \ER(\hat \Phi, \hat F^{(H+1)})} \\
    \lesssim\;& \calC^{(H+1)} \sigma_z \sqrt{\log(T)} \sqrt{\frac{k n_x \log\paren{N_1 T \frac{\bar \lambda}{\underline{\lambda}}} }{ cN_1 T H} + \frac{kn_u +\log(\frac{1}{\delta'})}{N_2 T}},
    \end{split}
\end{align}
where
\begin{align*}
    \calC^{(H+1)} &= \calJ(A_{\mathsf{cl}})\norm{B} \paren{\lambda_{\max} (Q)^{1/2} + \lambda_{\max}(R)^{1/2} \paren{\norm{K^{(H+1)}} + \frac{1}{2\calJ(A_{\mathsf{cl}})\norm{B}} }} \\
    &\quad\quad+ \lambda_{\max}(R)^{1/2} \sqrt{\frac{\trace\paren{\Sigma_x^{(H+1)}}}{\lambda_{\min}\paren{\Sigma_x^{(H+1)}}} } \\
    &\cong \lambda_{\max} (Q)^{1/2} \calJ(A_{\mathsf{cl}})\norm{B} + \lambda_{\max}(R)^{1/2} \paren{\norm{K^{(H+1)}} + \sqrt{\frac{\trace\paren{\Sigma_x^{(H+1)}}}{\lambda_{\min}\paren{\Sigma_x^{(H+1)}}} } }.
\end{align*}
This completes the proof.
\end{proof}

\end{document}